\PassOptionsToPackage{table,xcdraw,dvipsnames}{xcolor}

\documentclass{article}

\PassOptionsToPackage{numbers, compress}{natbib}

\usepackage[preprint]{neurips_2025}

\usepackage[utf8]{inputenc} 
\usepackage[T1]{fontenc}    
\usepackage{hyperref}       
\hypersetup{
    colorlinks=true,
    linkcolor=BlueViolet,
    citecolor=teal
}
\usepackage{url}            
\usepackage{booktabs}       
\usepackage{amsfonts}       
\usepackage{nicefrac}       
\usepackage{microtype}      
\usepackage{xcolor}         
\usepackage{ulem}
\usepackage{makecell}
\usepackage{amsthm}
\usepackage{bm}
\usepackage[table,xcdraw]{xcolor}
\usepackage{multicol}
\usepackage{colortbl}
\usepackage{amsmath}
\usepackage{cleveref}
\usepackage{multirow}
\usepackage{graphicx}
\usepackage{subcaption}
\usepackage{enumitem}
\usepackage{bbding}
\usepackage{color}

\usepackage{algorithm}
\usepackage{listings}
\usepackage{natbib}
\usepackage{tcolorbox}

\usepackage{mdframed}
\usepackage{amsmath,bm,bbm}
\usepackage{footnote}
\usepackage{adjustbox}

\usepackage{epigraph}
\definecolor{baselinecolor}{gray}{.9}

\usepackage{etoolbox}
\makeatletter
\AfterEndEnvironment{algorithm}{\let\@algcomment\relax}
\AtEndEnvironment{algorithm}{\kern2pt\hrule\relax\vskip3pt\@algcomment}
\let\@algcomment\relax
\newcommand\algcomment[1]{\def\@algcomment{\footnotesize#1}}
\renewcommand\fs@ruled{\def\@fs@cfont{\bfseries}\let\@fs@capt\floatc@ruled
  \def\@fs@pre{\hrule height.8pt depth0pt \kern2pt}%
  \def\@fs@post{}%
  \def\@fs@mid{\kern2pt\hrule\kern2pt}%
  \let\@fs@iftopcapt\iftrue}
\makeatother
\definecolor{tabhighlight}{HTML}{e5e5e5}
\definecolor{tabhighlight2}{HTML}{e8e8e8}

\newcommand{\para}[1]{
  \noindent\textbf{#1}
}
\newcommand{\red}[1]{
  \color{red}{#1}\color{black}
}
\usepackage{listings}
\definecolor{ForestGreen}{RGB}{34,139,34}

\renewcommand{\paragraph}[1]{\medskip\noindent\textbf{#1.~}}

\theoremstyle{plain}
\newmdtheoremenv{corollary}{Corollary}

\newmdtheoremenv[linewidth=0pt,innerleftmargin=4pt,innerrightmargin=4pt]{lemma}{Lemma}%

\usepackage{tikz}

\newcommand*{\circled}[1]{\lower.7ex\hbox{\tikz\draw (0pt, 0pt)%
    circle (.5em) node {\makebox[1em][c]{\small #1}};}}

\title{Coding Triangle: How Does Large Language Model Understand Code?}

\author{Taolin Zhang$^{1,2,*}$, Zihan Ma $^{1,3,*}$, Maosong Cao$^{1}$, Junnan Liu$^{1}$,\\ \textbf{ Songyang Zhang$^{1,\dagger,\ddagger}$, Kai Chen$^{1,\dagger}$}\\
$^1$Shanghai AI Laboratory $^2$Tsinghua University $^3$Xi’an Jiaotong University \\
}

\begin{document}

\maketitle
\begin{abstract}
Large language models (LLMs) have achieved remarkable progress in code generation, yet their true programming competence remains underexplored. We introduce the \textbf{Code Triangle} framework, which systematically evaluates LLMs across three fundamental dimensions: editorial analysis, code implementation, and test case generation. Through extensive experiments on competitive programming benchmarks, we reveal that while LLMs can form a self-consistent system across these dimensions, their solutions often lack the diversity and robustness of human programmers. We identify a significant distribution shift between model cognition and human expertise, with model errors tending to cluster due to training data biases and limited reasoning transfer. Our study demonstrates that incorporating human-generated editorials, solutions, and diverse test cases, as well as leveraging model mixtures, can substantially enhance both the performance and robustness of LLMs. Furthermore, we reveal both the consistency and inconsistency in the cognition of LLMs that may facilitate self-reflection and self-improvement, providing a potential direction for developing more powerful coding models. \footnote{This work is done when Taolin Zhang and Zihan Ma
are on internship at Shanghai AI Laboratory, * means equal
contribution, $^{\dagger}$ means corresponding author, $^{\ddagger}$ means project
lead.}
\end{abstract}

\section{Introduction}
\label{sec: introduction}
Recent advances in large language models (LLMs) with rapid development in model design and data scaling \cite{gpt3, gpt4, o3, llama2, llama3, mistral, qwen, qwen2, qwen3, claude3.5, gpt4o} have driven remarkable progress on code generation benchmarks \cite{chen2021evaluatinglargelanguagemodels, austin2021programsynthesislargelanguage, cassano2022multiplescalableextensibleapproach, jain2024livecodebench,m2rceval,autokaggle,codeeditorbench,tablebench}. For example, DeepSeek-V3 \cite{liu2024deepseek} achieves a score of 65.2 on HumanEval \cite{guo2025deepseek, chen2021evaluatinglargelanguagemodels}, Qwen3 attains 65.7 on LiveCodeBench \cite{qwen3, jain2024livecodebench}, and o3 reaches a Codeforces Elo rating of 2724 \cite{o3, el2025competitive}, demonstrating impressive coding abilities of modern foundation models.

With rapid advancements in coding capabilities of large language models, a growing concern is that current coding benchmarks fail to accurately and comprehensively evaluate the coding ability of LLMs. In this work, we address a foundational question: \textbf{How should the coding capability of LLM be defined?} When human developers solve coding problems, they typically follow a structured pipeline involving {problem analysis}, {designing a preliminary solution strategy}, {code implementation}, {manual testing}, and {iterative refinement} until the solution passes all test cases.

Motivated by this, we investigate the coding capabilities of LLMs through three interconnected dimensions: \textbf{Editorial}, \textbf{Code}, and \textbf{Cases}. We develop such a three-dimensional analysis framework called \textbf{Coding Triangle} to study the coding behaviors of LLMs.  Our goal is to understand how LLMs fundamentally interpret coding problems within each dimension and to explore the interactions across them. For example: \textit{(a) Does performance across dimensions exhibit consistency?}\textit{(b) Does a model’s code generation benefit from its natural language problem breakdowns?}\textit{(c) Does the generated code reliably pass self-generated test cases?}\textit{(d) To what extent do these test cases adequately reflect the reasoning outlined in the editorial?}
Based on our Coding Triangle framework, we conduct extensive experiments on 200 problems collected from AtCoder and evaluate various LLMs including general models, coding model and reasoning model. By analyzing their capabilities and interactions among different dimensions, we gain deeper insights into how LLMs truly comprehend and tackle coding tasks with several surprising and interesting findings.

Our results demonstrate prevalent \textbf{self-consistency} across three dimensions in various LLMs. This self-consistency often confines LLM reasoning, leading it to converge on narrow patterns and repeat similar errors, particularly with corner cases or implementation details. Consequently, a significant distribution shift emerges between LLM predictions and human submissions. Notably, ensembling multiple models can effectively mitigate these cognitive biases, enhancing performance diversity and robustness.
Conversely, we observe \textbf{self-inconsistency} across these dimensions, indicating that their respective capabilities are not always fully aligned. An LLM might, for example, accurately analyze its own failed solutions to pinpoint root causes and effectively use self-generated test cases to differentiate correct from incorrect solutions; yet, these individual strengths may not be cohesively integrated, exemplifying the aforementioned misalignment. These findings suggest potential for self-reflection and self-improvement by iteratively aligning these dimensions.

To summarize, this paper makes the following contributions.

\begin{figure}[!t]
    \centering    \includegraphics[width=\textwidth, page=3]{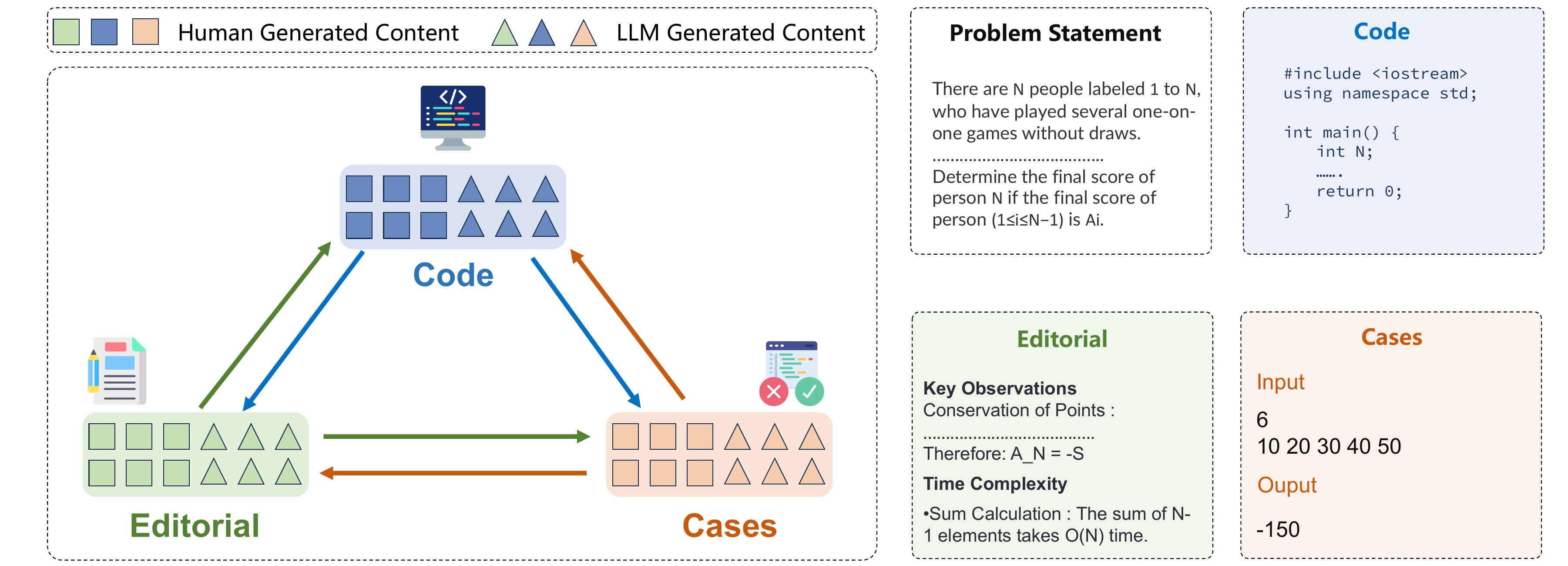}
    \vspace{-5pt}
    \caption{
        \small \textbf{The framework of Coding Triangle.} Editorial, code, and cases form the three fundamental vertices of the triangle, with each vertex can be sampled from either human solutions or model predictions. These vertices are interconnected, influencing one another, and their relationships form the six directed edges of the triangle, representing the mutual interactions between Editorial, Code, and Cases.
    }
    \label{fig:eq}
\end{figure}



\begin{itemize}[leftmargin=1.5em]
        \item[$\bullet$] We propose a framework called \textbf{Coding Triangle} to systematically examine the internal knowledge of LLMs in programming, enabling a comprehensive evaluation of their coding abilities.
        \item[$\bullet$] We investigate the distribution shift between the LLM predictions  and actual solutions from human, and find that incorporating human information can substantially improve performance.
        \item[$\bullet$] By analyzing the self-consistency and self-inconsistency inside LLMs, we observe that their strengths and weaknesses vary across the three dimensions in Coding Triangle, demonstrating the advantages of model mixtures and the potential for self-reflection and self-improvement.
    \end{itemize}

\section{Overview}

In this section, we first introduce the three fundamental vertices of Coding Triangle and define how to evaluate the ability of LLM across these three dimensions. We will proceed with the analysis and explore the distribution shift compared to the distribution of human coding in Section \ref{sec:method}, and discuss the interaction among these three dimensions in Section \ref{sec:edge}. 
Our experiments cover general models, coder models and reasoning models\footnote{We denote the DeepSeek-V3 \cite{liu2024deepseek}, Qwen2.5-72B-Instruct\cite{yang2024qwen2}, Qwen2.5-Coder-32B-Instruct \cite{hui2024qwen2} and QwQ-32B \cite{qwq32b}, as DS-V3, 72B, Coder and QWQ for short, respectively. All the results are obtained with Nvidia-A800. } and we utilize  AtCoder\footnote{Problems A\text{-}F from AtCoder Beginner Contest (ABC) 175\text{-}374.} as the evaluation problem sets.


\subsection{Coding Triangle}
In this study, we systematically examine the comprehensive understanding of LLMs when addressing competitive programming challenges. 
Motivated by how human solve the problem steps by steps through analyzing, coding and manually testing, we formally decompose coding ability into three interconnected perspectives, and establish the overall framework of \textbf{{Coding Triangle}}:

\noindent$\bullet$ \textbf{\textit{Editorial}} captures how LLM interprets and analyzes the  the problem in natural language, providing the most accessible explanation for human readers.

\noindent$\bullet$  \textbf{\textit{Code}} reflects the programming logic and ability of algorithm implementation, serving as the machine-executable counterpart to the human-readable editorial.

\noindent$\bullet$  \textbf{\textit{Cases}} indicate the depth of understanding in terms of validation criteria, including edge scenarios and boundary conditions.

Intuitively, these three dimensions create a comprehensive system that captures all aspects of a coding problem, from interpretation to execution and validation. We then introduce evaluation metrics for each dimension, allowing us to quantify its strengths and weaknesses in a structured way.

\subsection{Evaluation Metric}
\para{Editorial.} 
We adopt an LLM-as-a-judge approach to evaluate the quality of model-generated editorials. With the model-generated editorial $E_{\mathrm{model}}$ and the ground-truth editorial $E_{\mathrm{gt}}$, we require o3-mini \cite{o3mini} to judge them and predict a correctness score. Let $N$ be the total number of problems, 
the overall editorial score $S_{\mathrm{edi}}$ is defined as
\begin{equation}
S_{\mathrm{edi}} = \frac{1}{N} \sum_{i=1}^{N} \mathrm{LLM}(E_{\mathrm{model}}^{(i)}, E_{\mathrm{gt}}^{(i)}).
\end{equation}

\para{Code.}
During the evaluation, the model is prompted to generate a solution for each problem, which is then validated against all public ground-truth test cases. For each problem $i$, let $\mathcal{T}_i$ denote the set of public ground-truth test cases, $\mathcal{J}$ as the judge function and $s_i$ the solution from model. With $N$ as the total number of problems, we count the number of problems where the solution passes all test cases:
$N_{\mathrm{code}} = \left| \left\{ i \mid \forall t \in \mathcal{T}_i,\; \mathcal{J}(s_i, t) = \mathrm{Accepted} \right\} \right|$
. The code score is defined as Pass@1 accuracy:
\begin{equation}
S_{\mathrm{code}} = \frac{N_{\mathrm{code}}}{N}.
\end{equation}

\para{Cases.}
It is observed that official cases for some problems are not sufficiently comprehensive and may fail to cover all edge scenarios, potentially leading to incorrect solutions being accepted. Therefore, we focus on those errors that can be identified by the official cases and evaluate model-generated cases with only incorrect solutions. Furthermore, directly generating input-output pairs \cite{chen2021evaluating, liu2023your} is often insufficient, as most generated cases are incorrect and will be filtered out. Instead, we prompt the model to generate 
inputs and use the official solution to produce the outputs.

For each problem $i$, let $\mathcal{H}_i^{\mathrm{wrong}}$ denote the set of all human submissions that are incorrect, and $\mathcal{J}(h, \mathcal{C}_i)$ 
as the judge results of submission $h$ on the model-generated cases $\mathcal{C}_i$. We consider the cases set to be correct if the judge results are consistent with the official evaluation and all wrong solutions are identified as incorrect. Let $N_{\mathrm{case}}$ to be the number of problems for which the model-generated cases can effectively distinguish all the edge cases, we have
$
N_{\mathrm{case}} = \left| \left\{ i \mid \forall h \in \mathcal{H}_i^{\mathrm{wrong}},\; \mathcal{J}(h, \mathcal{C}_i) = \mathcal{J}(h, \mathcal{T}_i) \right\} \right|$
,
where $\mathcal{T}_i$ is the set of ground-truth test cases for problem $i$. The case score is then defined as
\begin{equation}
S_{\mathrm{case}} = \frac{N_{\mathrm{case}}}{N}.
\end{equation}
\section{Ability Analysis and Distribution Shift}
\label{sec:method}
With the evaluation metrics established above, we analyze the capabilities of LLMs across the three dimensions in this section. Furthermore, we explore the distribution shift between model cognition and human submissions, and demonstrate the robustness introduced by model mixture.

\begin{figure*}[!t]
	\centering
	\begin{subfigure}{0.24\linewidth}
		\includegraphics[width=\linewidth]{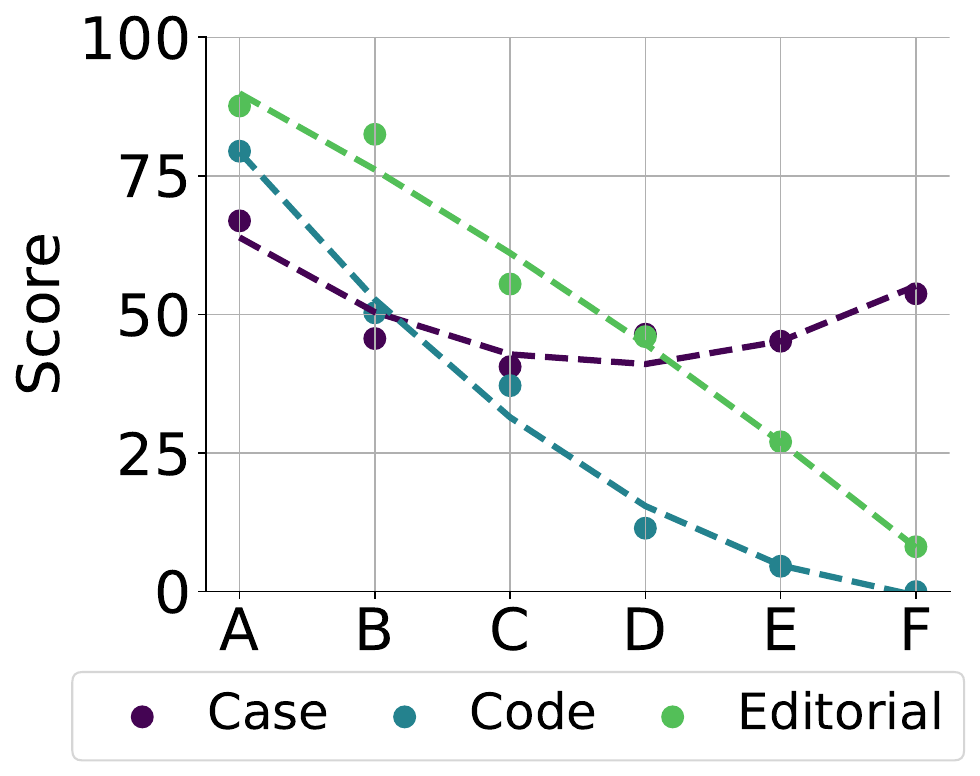}
		\caption{Qwen2.5-72B-Instruct}
		\label{fig:qwen_72b_stats}
	\end{subfigure}
    \hfill
	\begin{subfigure}{0.24\linewidth}
		\includegraphics[width=\linewidth]{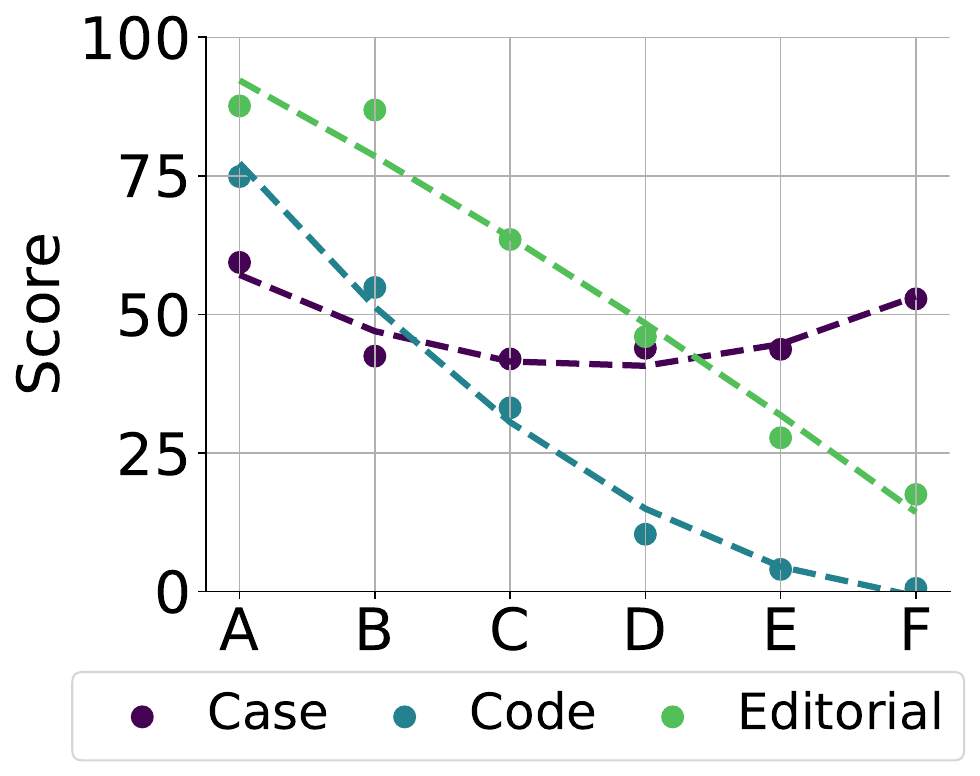}
		\caption{Qwen2.5-Coder-32B}
		\label{fig:coder_32b_stats}
	\end{subfigure}
    \hfill
    \begin{subfigure}{0.24\linewidth}
        \includegraphics[width=\linewidth]{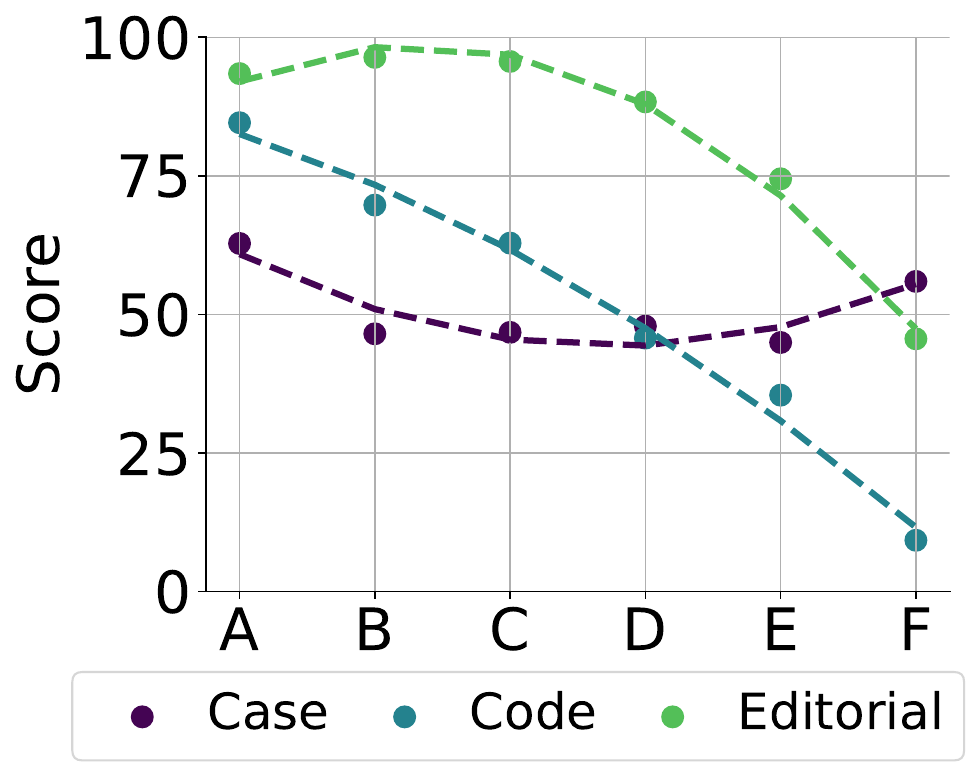}
        \caption{DeepSeek-V3}
        \label{fig:v3_stats}
    \end{subfigure}
    \hfill
	\begin{subfigure}{0.24\linewidth}
		\includegraphics[width=\linewidth]{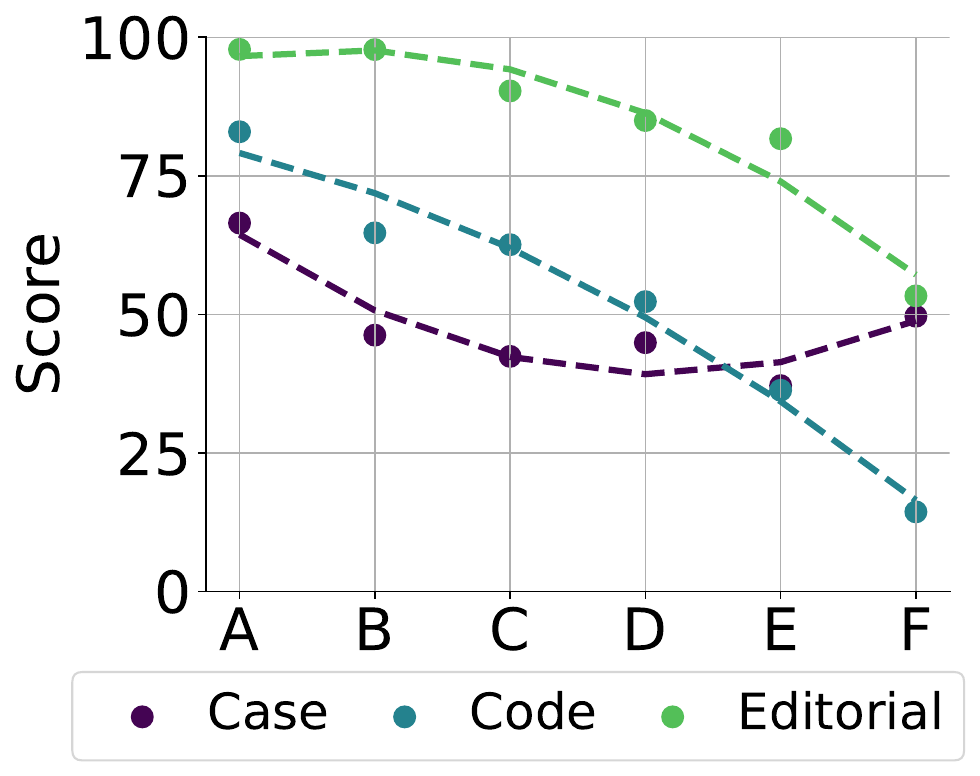}
		\caption{QwQ-32B}
		\label{fig:qwq_32b_stats}
	\end{subfigure}
	\caption{
    Ability analysis across different dimensions.
    }
	\label{fig:all_traingle}
\end{figure*}

\subsection{Ability Analysis}
We present the performance across the three dimensions in Figure~\ref{fig:all_traingle} to illustrate their relationships. It can be observed that models with strong coding abilities tend to perform well across the editorial dimension and the code dimension, indicating a consistency between these two dimensions. As problem difficulty increases, both the editorial analysis and code generation abilities decline. Notably, the editorial score remains consistently higher than the code score, resulting in a performance gap that typically ranges from 0\% to 20\% when the model translates its problem analysis into executable code. This trend is observed across all models, including reasoning-oriented ones such as QwQ-32B, where correct reasoning does not always lead to correct solutions. Moreover, this gap is most pronounced on medium-difficulty problems, suggesting the conflict between reasoning and implementation. 

In contrast, the case score exhibits a different trend and does not decrease monotonically as problem difficulty increases. And we surprisingly find that the code score can even surpass editorial and code score on the most difficult problems. Interestingly, performance drops on medium-difficulty problems, as seen in problems C and D, which we attribute to the increased complexity and abundance of edge cases at this level. For harder problems such as E and F, the main challenge shifts to algorithmic selection and advanced techniques rather than numerous edge conditions, resulting in a rebound in case accuracy. These findings suggest that the ability to generate test cases does not fully align with editorial and coding abilities, indicating an inconsistency among the three dimensions.

\subsection{Distribution Shift}
In practice, LLMs are capable of performing problem analysis, code generation, and test case generation to form a self-consistent system. However, this system is actually limited to its own cognition, and has a significant distribution shift from human solutions. In the following part, we explore how the  self-cognition causes this distribution shift. Since editorial evaluation is based on subjective assessment using LLM-as-Judge, we only investigate the distribution shift results in objective evaluations of code and test cases.

\begin{figure*}[!t]
	\centering
	\begin{subfigure}{0.24\linewidth}
		\includegraphics[width=\linewidth]{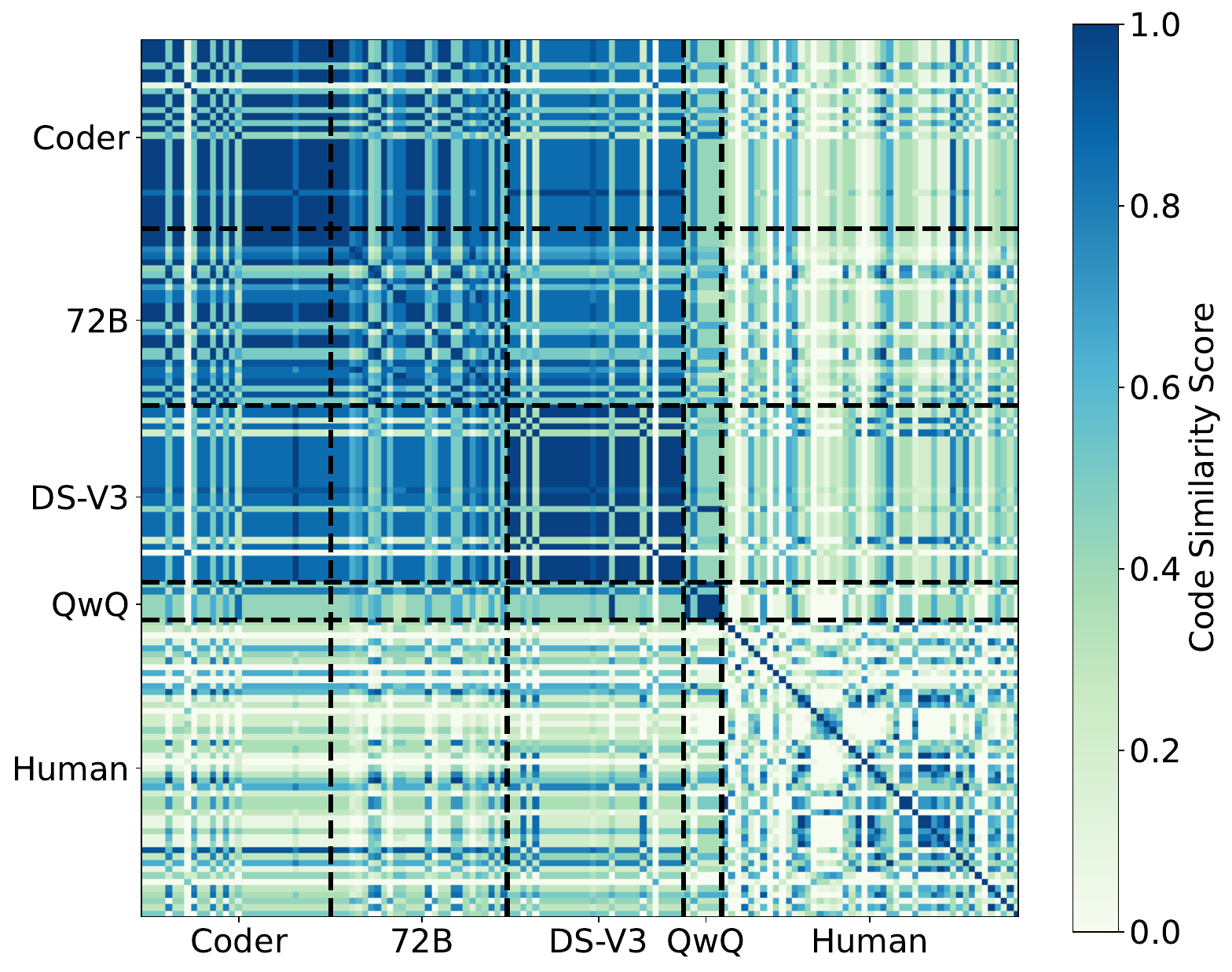}
		\caption{\small Code Similarity Score.}
		\label{fig:code_heatmap}
	\end{subfigure}
	\hfill
	\begin{subfigure}{0.24\linewidth}
		\includegraphics[width=\linewidth]{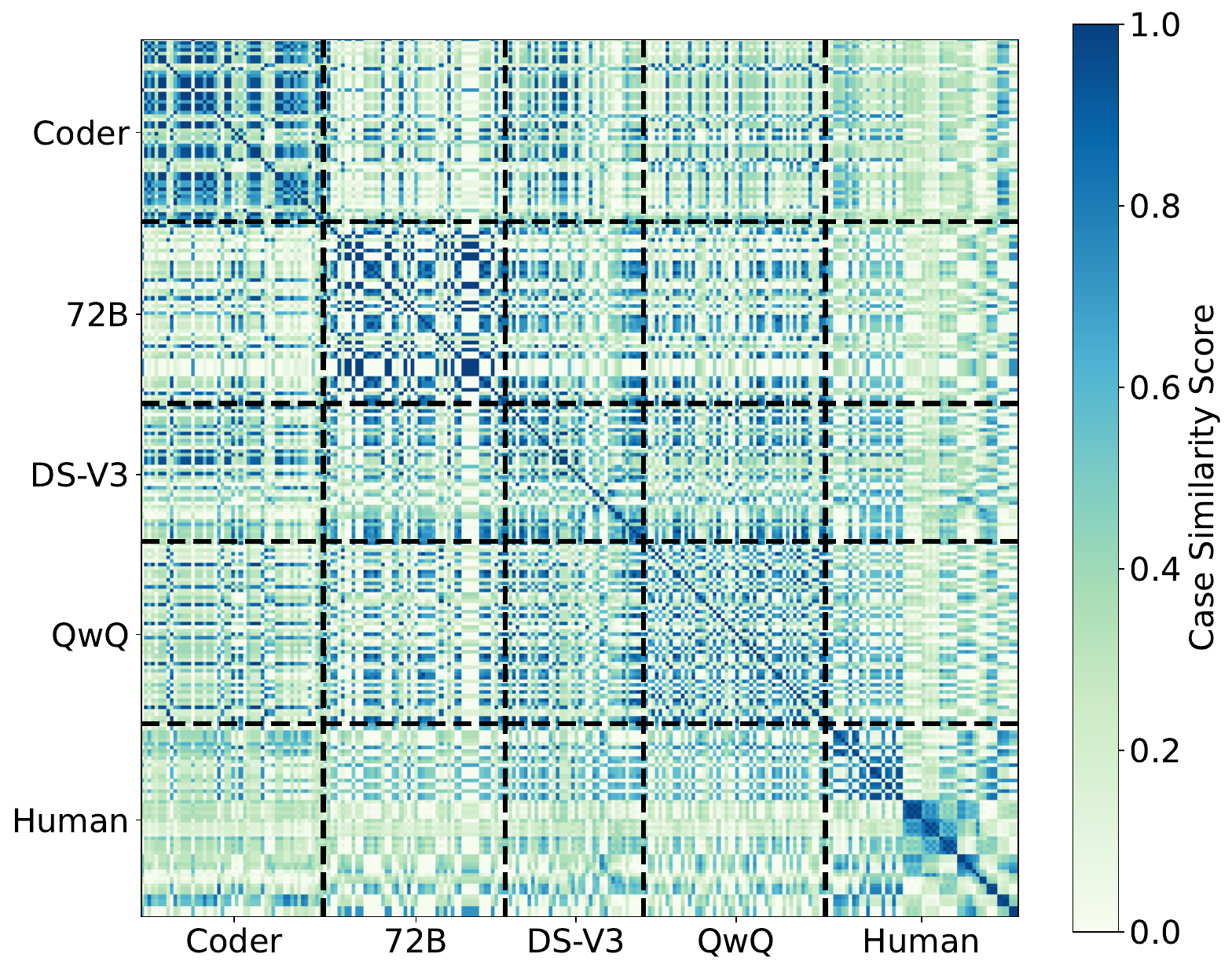}
		\caption{\small Case Similarity Score. }
		\label{fig:case_heatmap}
	\end{subfigure}
	\begin{subfigure}{0.24\linewidth}
		\includegraphics[width=\linewidth]{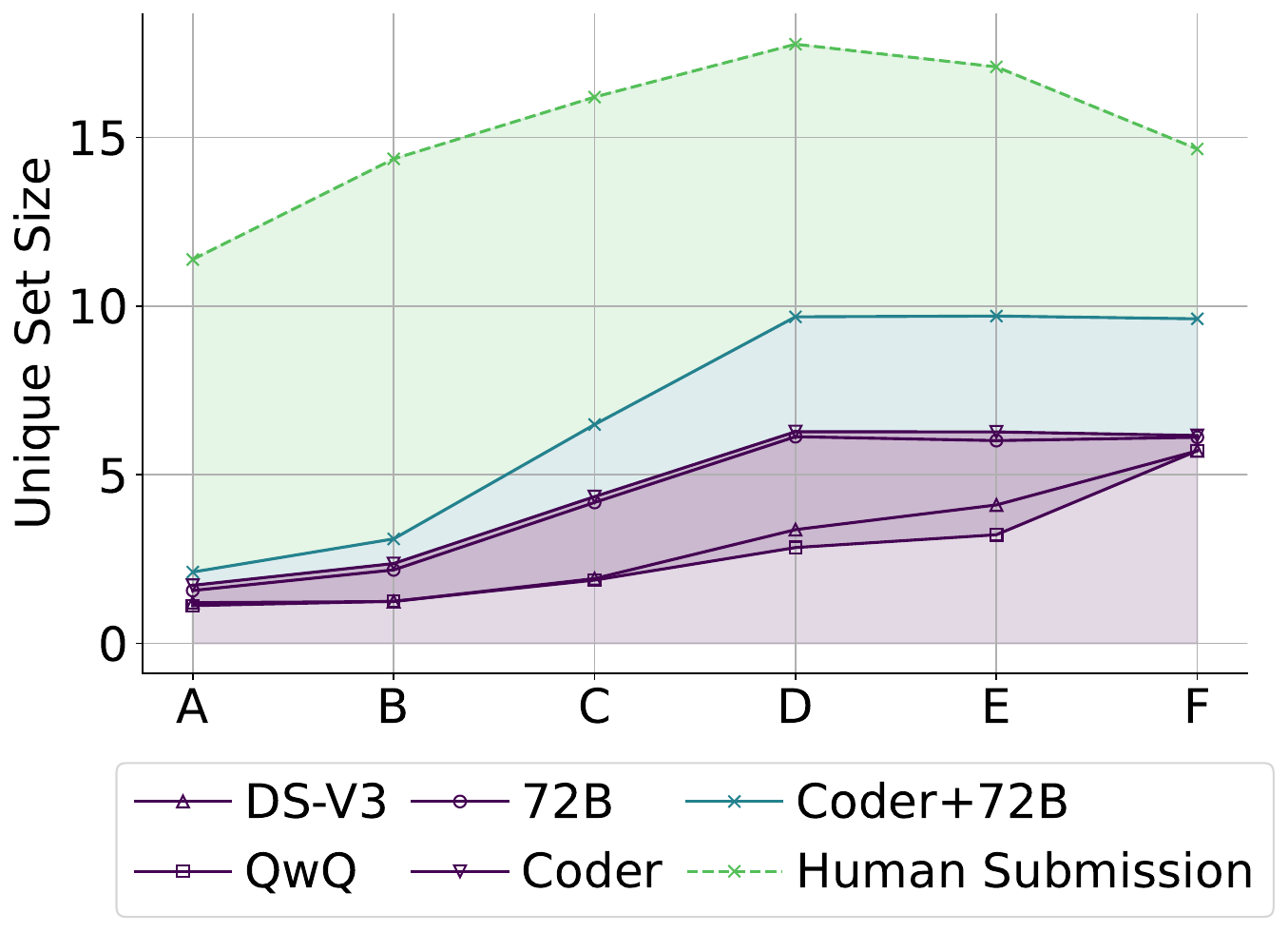}
		\caption{\small Code Unique Set Size.}
		\label{fig:code_unique}
	\end{subfigure}
	\hfill
	\begin{subfigure}{0.24\linewidth}
		\includegraphics[width=\linewidth]{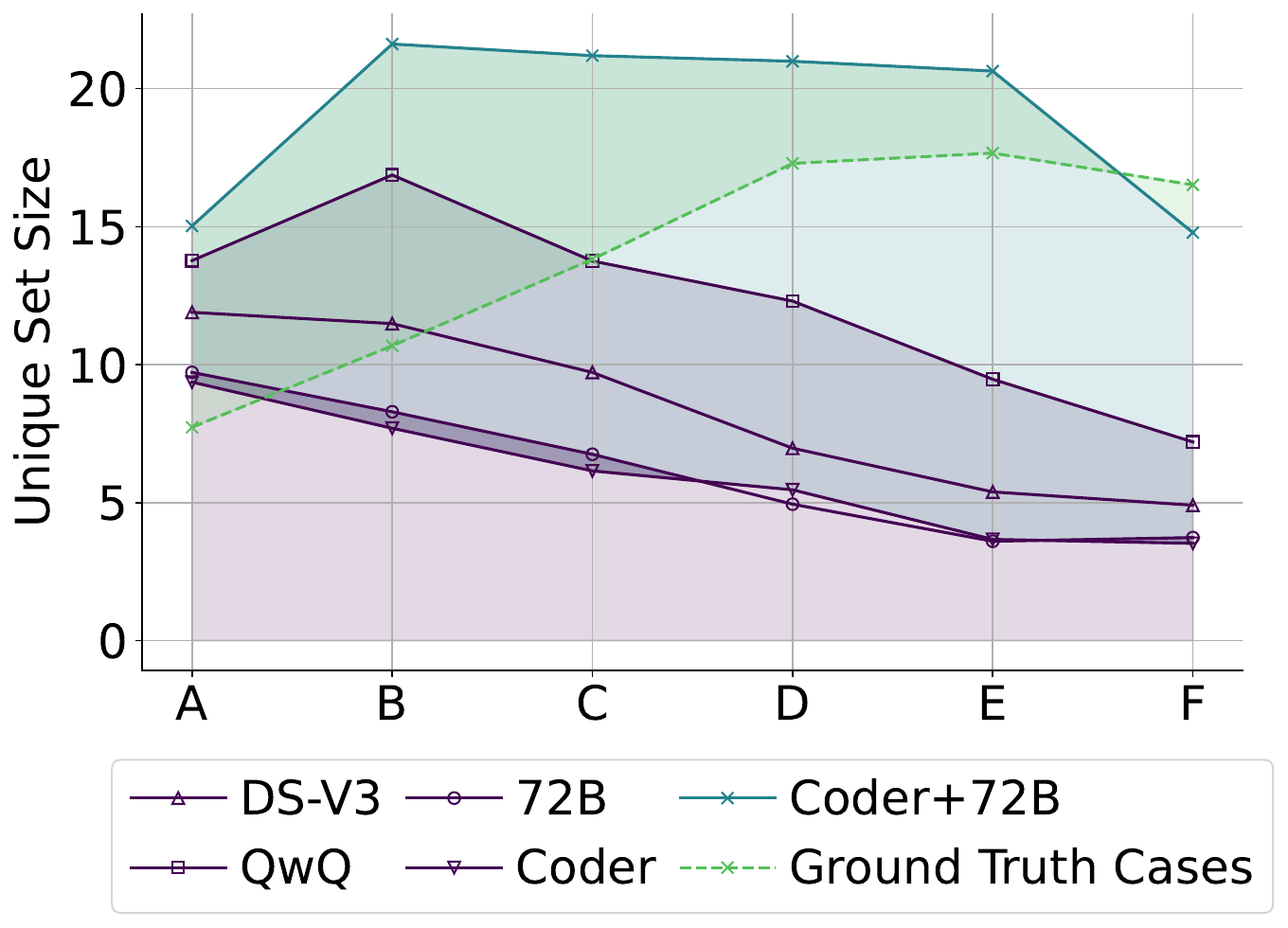}
		\caption{\small Cases Unique Set Size.}
		\label{fig:case_unique}
	\end{subfigure}
	\caption{Similarity analysis and unique set size for error analysis of codes and cases.}
\end{figure*}

\para{Distribution Shift on Code.}
To analyze the distribution shift between model and human solutions, we first construct a performance matrix $P_{\text{code}} \in \mathbb{R}^{m \times n}$ for each problem, where $m$ is the total number of solutions (including those generated by different models and those submitted by humans), and $n$ is the number of test cases. Each entry $P_{ij}$ is assigned a value of $1$ if solution $i$ passes test case $j$, and $-1$ otherwise.
For error analysis, we exclude solutions that pass all test cases.
Each row of $P_{\text{code}}$ represents the performance vector of a specific solution. We normalize these vectors and compute the cosine similarity between every pair of solutions:
\begin{equation}
\text{sim}(i, k) = \frac{P_{i} \cdot P_k}{|P_i| |P_k|}, \quad \forall i, k \in {1, \dots, m}
\end{equation}
The resulting similarity matrix is visualized in Figure~\ref{fig:code_heatmap}. Our results reveal that model-generated solutions are highly similar to each other, with most pairs exhibiting a similarity score above $0.8$. This suggests that model tends to make similar mistakes and generating multiple roll-outs does not prevent the inherent reasoning patterns. Notably, solutions produced by the same model show even higher internal cognition, while human-submitted solutions are much more diverse, displaying lower similarity scores and a wider variety of errors.

To further quantify the diversity of solutions, we construct a set of unique performance vectors for each problem and then compute the average size of the set, as illustrated in Figure~\ref{fig:code_unique}. Our findings reveal that human submissions display significantly greater diversity compared to model-generated solutions, indicating a wider variety of errors in human attempts. As problem difficulty increases, the diversity of errors in both model and human solutions grows, reflecting the heightened complexity of the tasks. Notably, combining solutions from multiple models yields a broader range of unique behaviors than relying on any single model, suggesting that different models produce distinct types of errors. By integrating solutions from various models, we can identify more boundary conditions, and a comprehensive analysis of the error encountered by these models may further enhance overall robustness and performance.

\para{Distribution Shift on Cases.}
We also conduct a similarity analysis on test cases and construct a performance matrix $P_{case}$ for each problem, where the rows correspond to human submissions and the columns correspond to test cases. We then use the column vectors of the matrix to represent the performance vector of each test case and calculate the cosine similarity between every pair of test cases, as shown in Figure~\ref{fig:case_heatmap}.
The results show that, unlike solutions, model-generated cases and ground-truth cases do not exhibit extremely high similarity. However, we still observe that cases generated by the same model, as well as ground-truth cases, tend to be more similar within their own groups. The reason for this is that both model-generated cases and official cases are often constructed following certain fixed patterns or templates, leading to internal redundancy.

We use the unique set as an analytical tool to evaluate the diversity of generated cases as well. Notably, for easier problems, the unique set size from the models exceeds that of the ground truth cases. Upon reviewing the official cases in these situations, we find that they tend to be relatively easy and short due to the simplicity of the problems. As a result, the model-generated cases are able to capture a wider range of potential errors. For harder problems, the unique set size decrease and the test cases from models fail to recognize different error of the solutions. 
Surprisingly, we observe that the combination of cases set generated from different models also shows larger unique set size, indicating mixture of different models provides a better generalization of cases.


\section{Bridging the Edges in Coding Triangle}
\label{sec:edge}
With the analysis framework established above, we now explore and quantify how the three dimensions of the coding triangle are interconnected, with a particular focus in the context of self-cognition inside LLMs. We also investigate how the introduction of external human knowledge, such as ground truth editorials, solutions, and test cases, affects these relationships. To this end, we systematically examine the six possible directed edges of the triangle, analyzing how providing information about one aspect (A), either from the model itself or from human sources, influences another aspect (X).

\subsection{From Editorial to X}
\para{\textit{From Editorial to Code: Does LLM benefit from the editorial generated by itself?}}
\label{sec:etc}

\begin{figure*}[!t]
	\centering
	\begin{subfigure}{0.24\linewidth}
		\includegraphics[width=\linewidth]{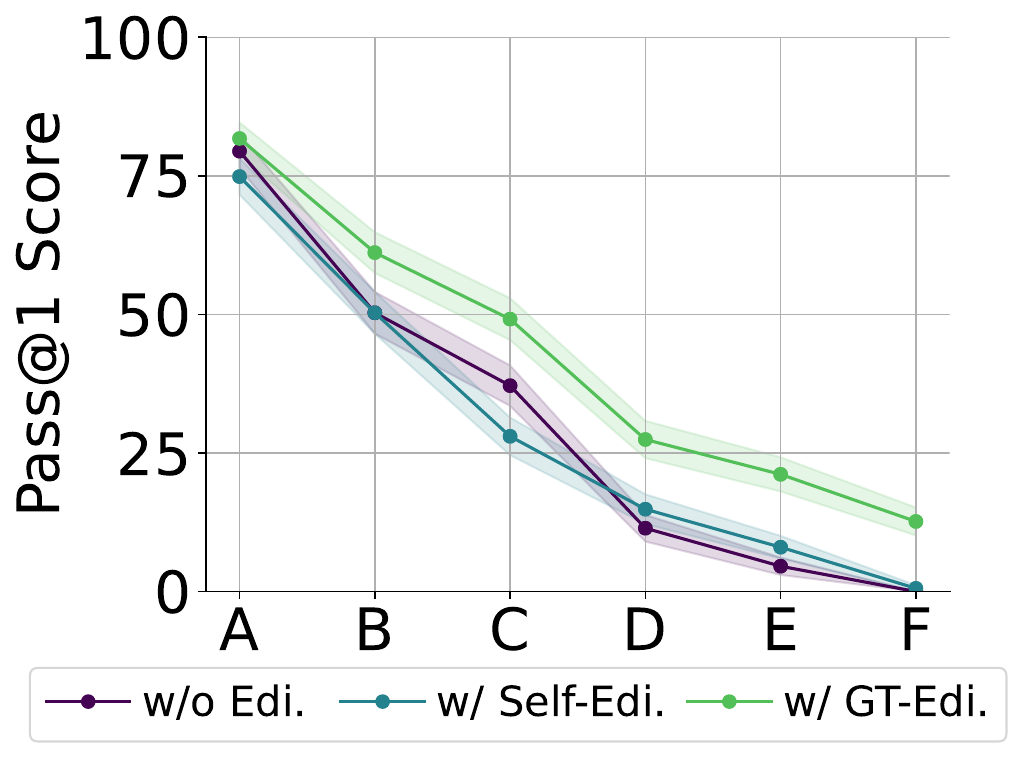}
		\caption{Qwen2.5-72B-Instruct}
		\label{fig:qwen_72b_etc}
	\end{subfigure}
    \hfill
	\begin{subfigure}{0.24\linewidth}
		\includegraphics[width=\linewidth]{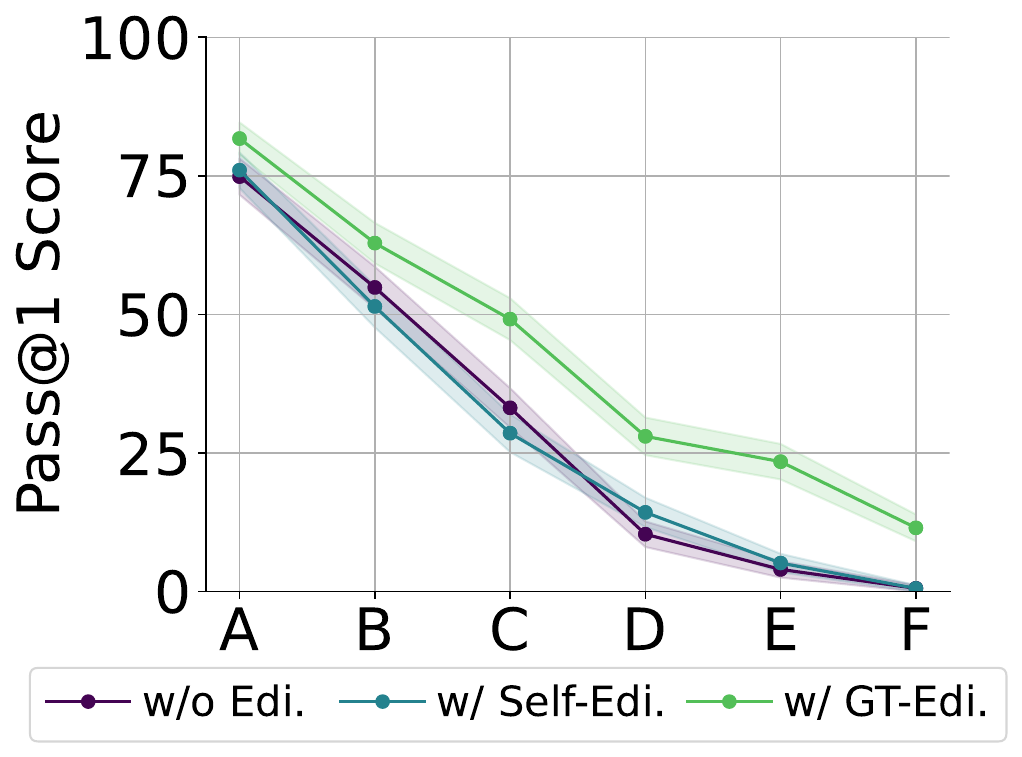}
		\caption{Coder-32B-Instruct}
		\label{fig:coder_32b_etc}
	\end{subfigure}
    \hfill
    \begin{subfigure}{0.24\linewidth}
        \includegraphics[width=\linewidth]{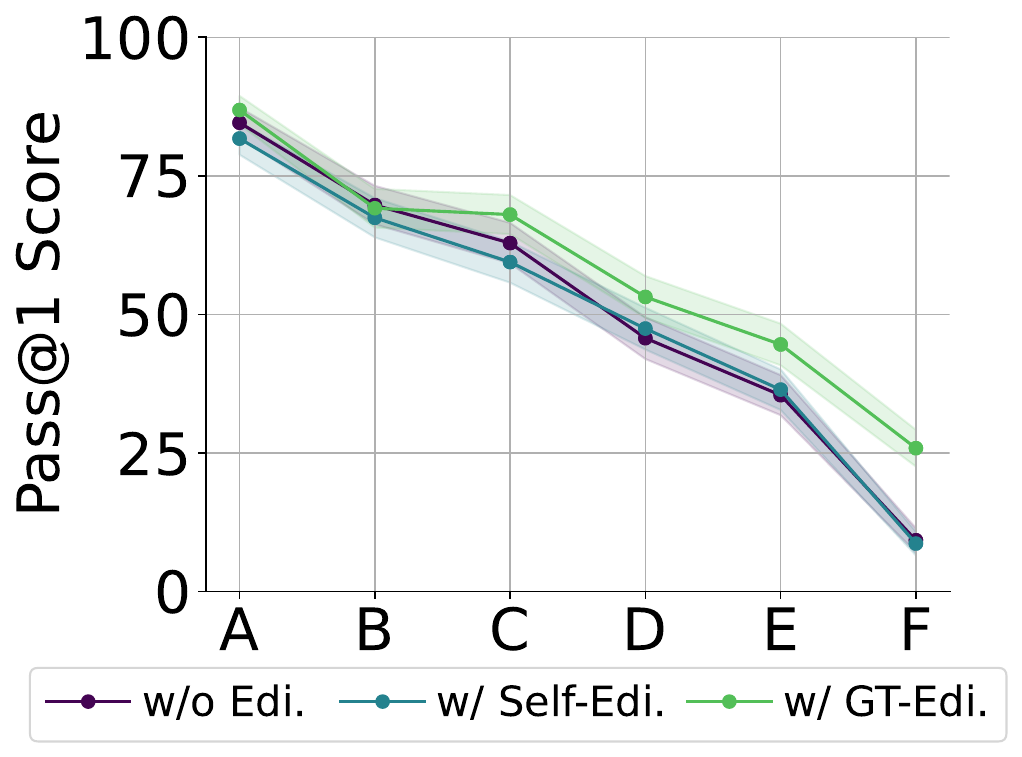}
        \caption{DeepSeek-V3}
        \label{fig:v3_etc}
    \end{subfigure}
    \hfill
	\begin{subfigure}{0.24\linewidth}
		\includegraphics[width=\linewidth]{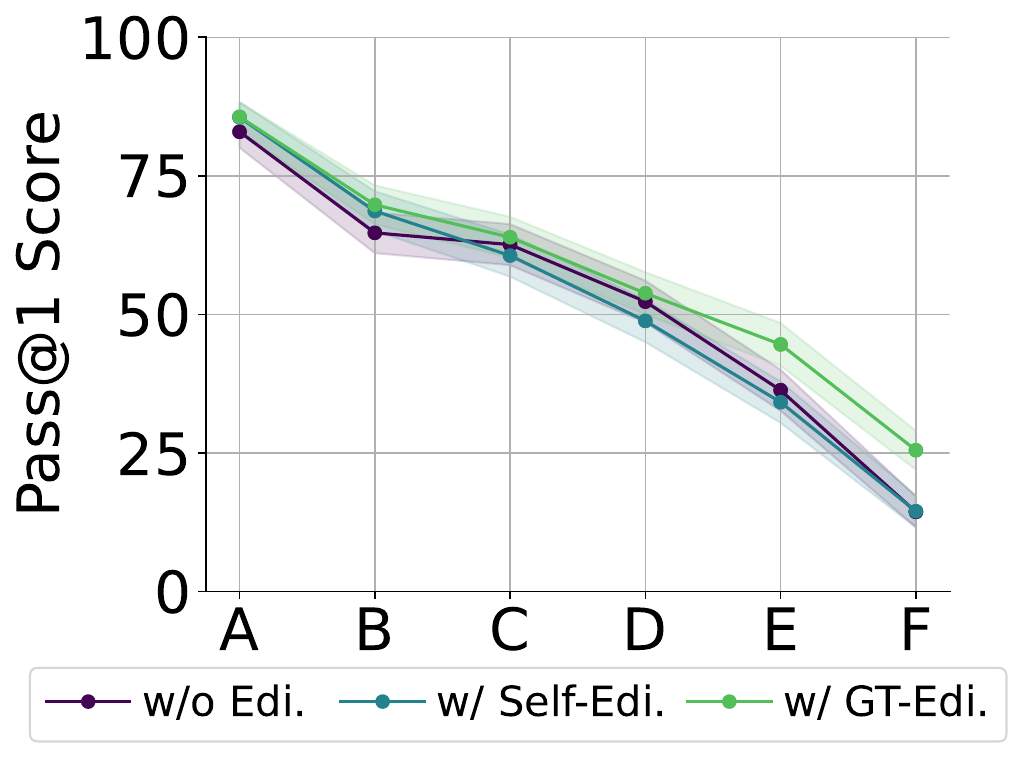}
		\caption{QwQ-32B}
		\label{fig:qwq_32b_etc}
	\end{subfigure}
	\caption{Pass@1 score with self-generated and ground truth editorials.}
	\label{fig:etc}
    \vspace{-0.5em}
\end{figure*}

To have a better understanding of the gap between the ability to analyze problems and  to implement code in LLM, we inspect the performance when feeding its own editorial to generate code, and study how its own problem understanding affects code implementation. For comparison, we also include the original pass rate as well as the results when feeding ground truth editorials, as shown in Figure~\ref{fig:etc}.

We observe that \textbf{providing self-generated editorials does not significantly enhance coding performance}. This suggests that the stage of problem analysis and code implementation are largely self-consistent. Even for reasoning-oriented models like QwQ, which perform reflection on both the editorial and the generated code, there is no notable performance boost. In some instances, prompting the model to write an editorial before coding can encourage CoT reasoning and slightly improve the pass rate. However, if the editorial contains flawed analysis, referencing it may actually reduce performance.
In contrast, providing ground truth editorials leads to a much more significant improvement in coding performance. However, even with access to ground truth editorials, model still fail to pass difficult problems, and high success rates are not always achieved, indicating the upper bound of its ability to utilize correct analysis for code generation.
Notably, we surprisingly find that \textbf{DS-V3 and QwQ display very similar patterns}. For easy problems, neither model gains much from ground truth editorials, suggesting that correct analysis alone is insufficient in that their internal reasoning and implementation details remain the primary bottlenecks. For harder problems, both models show comparable performance when prompted with ground truth editorials, implying that the main factor influencing the original pass rate is the their reasoning and understanding of the problem, rather than their code generation capability.

\begin{figure*}[!t]
	\centering
	\begin{subfigure}{0.24\linewidth}
		\includegraphics[width=\linewidth]{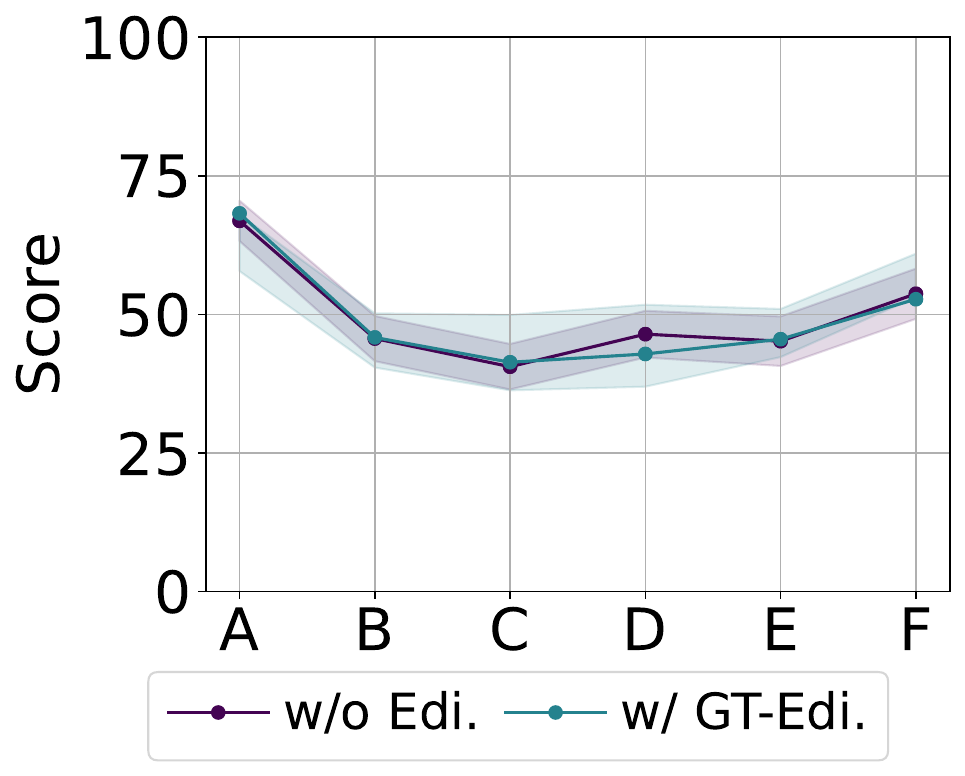}
		\caption{Qwen2.5-72B-Instruct}
	\end{subfigure}
    \hfill
	\begin{subfigure}{0.24\linewidth}
		\includegraphics[width=\linewidth]{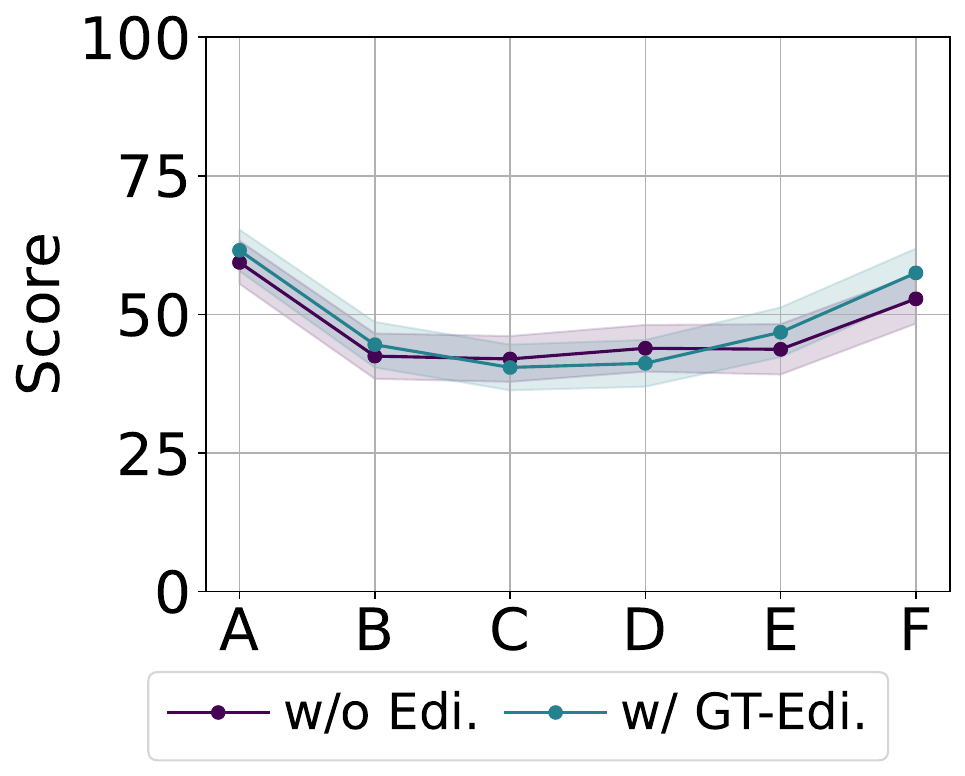}
		\caption{Coder-32B-Instruct}
	\end{subfigure}
    \hfill
    \begin{subfigure}{0.24\linewidth}
        \includegraphics[width=\linewidth]{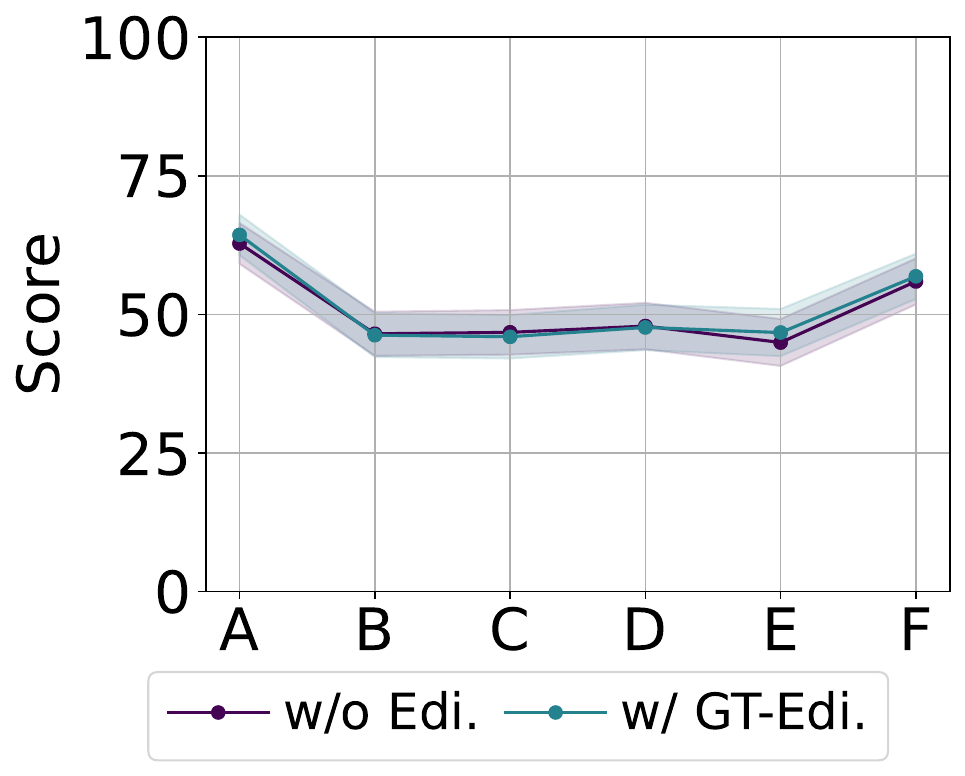}
        \caption{DeepSeek-V3}
    \end{subfigure}
    \hfill
	\begin{subfigure}{0.24\linewidth}
		\includegraphics[width=\linewidth]{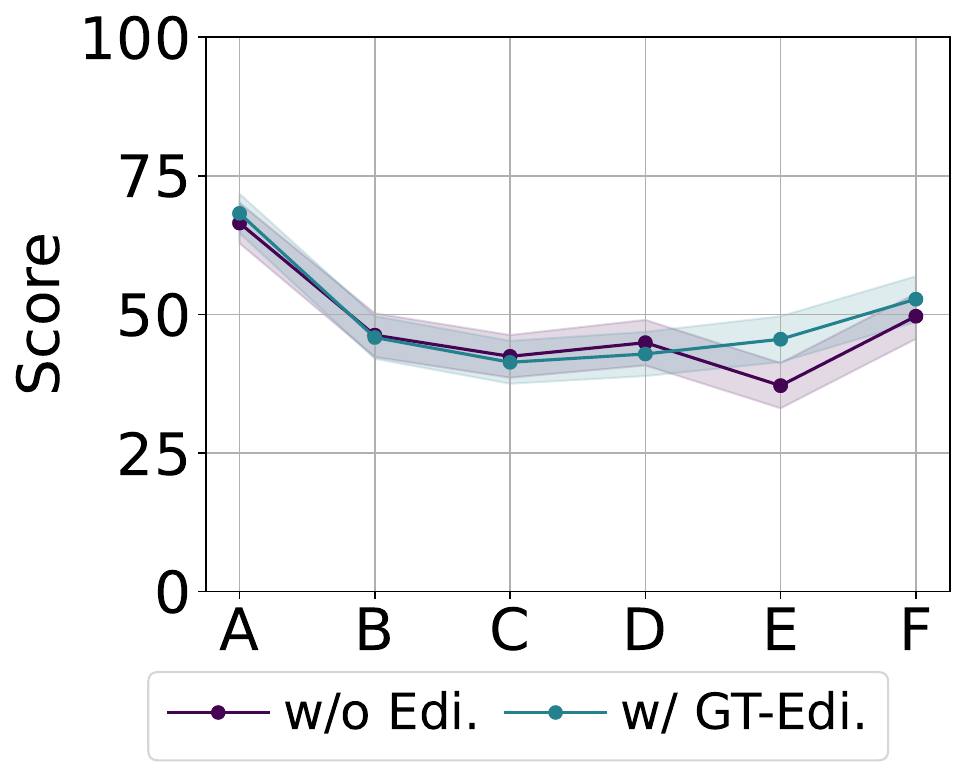}
		\caption{QwQ-32B}
	\end{subfigure}
	\caption{Case score w.r.t. ground truth editorials.}
	\label{fig:ett}
\end{figure*}
\para{\textit{From Editorial to Cases: Does case generation benefit from the ground truth editorials?}}

We further investigate the impact of editorials on test case generation. As shown in Figure~\ref{fig:ett}, \textbf{even when the model is provided with detailed human-written editorials, its ability to generate high-quality test cases does not significantly improve}. This suggests that the skills required for creating diverse and comprehensive test cases are distinct from those needed for code generation, and such skills are not effectively transferred through editorial.
In particular, test case creation is more closely linked to the specific implementation details of code, while editorials provide only high-level abstraction and are less effective in guiding the model to cover all edge scenarios. Therefore, simply providing high quality editorials is not sufficient to improve the capability in generating robust test cases, highlighting the necessity for other strategies to address this challenge.

\begin{figure*}[!t]
    \centering
    \begin{subfigure}{0.45\linewidth}
        \includegraphics[width=\linewidth]{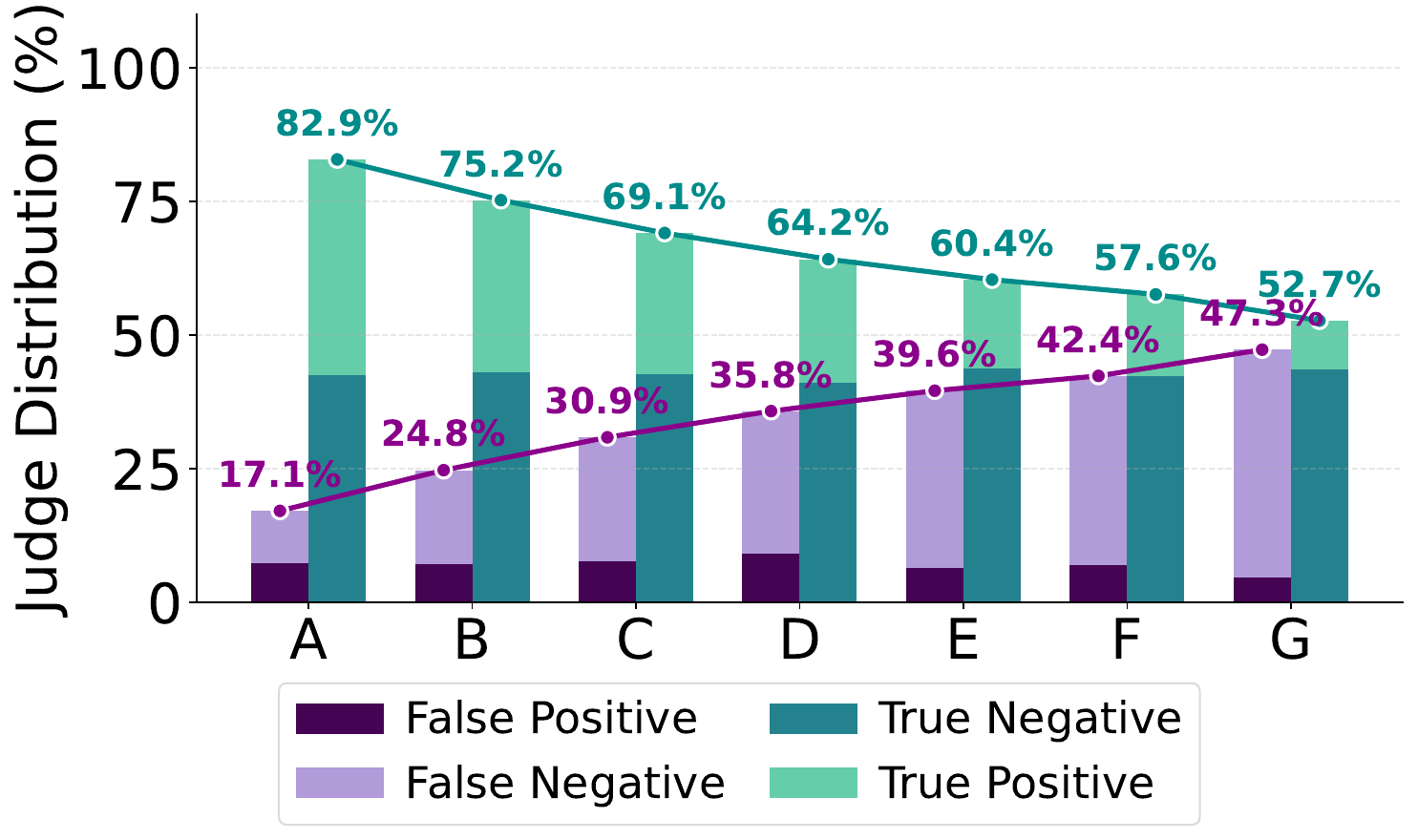}
        \caption{DeepSeek-V3 over human solutions.}
    \end{subfigure}
    \hfill
    \begin{subfigure}{0.45\linewidth}
        \includegraphics[width=\linewidth]{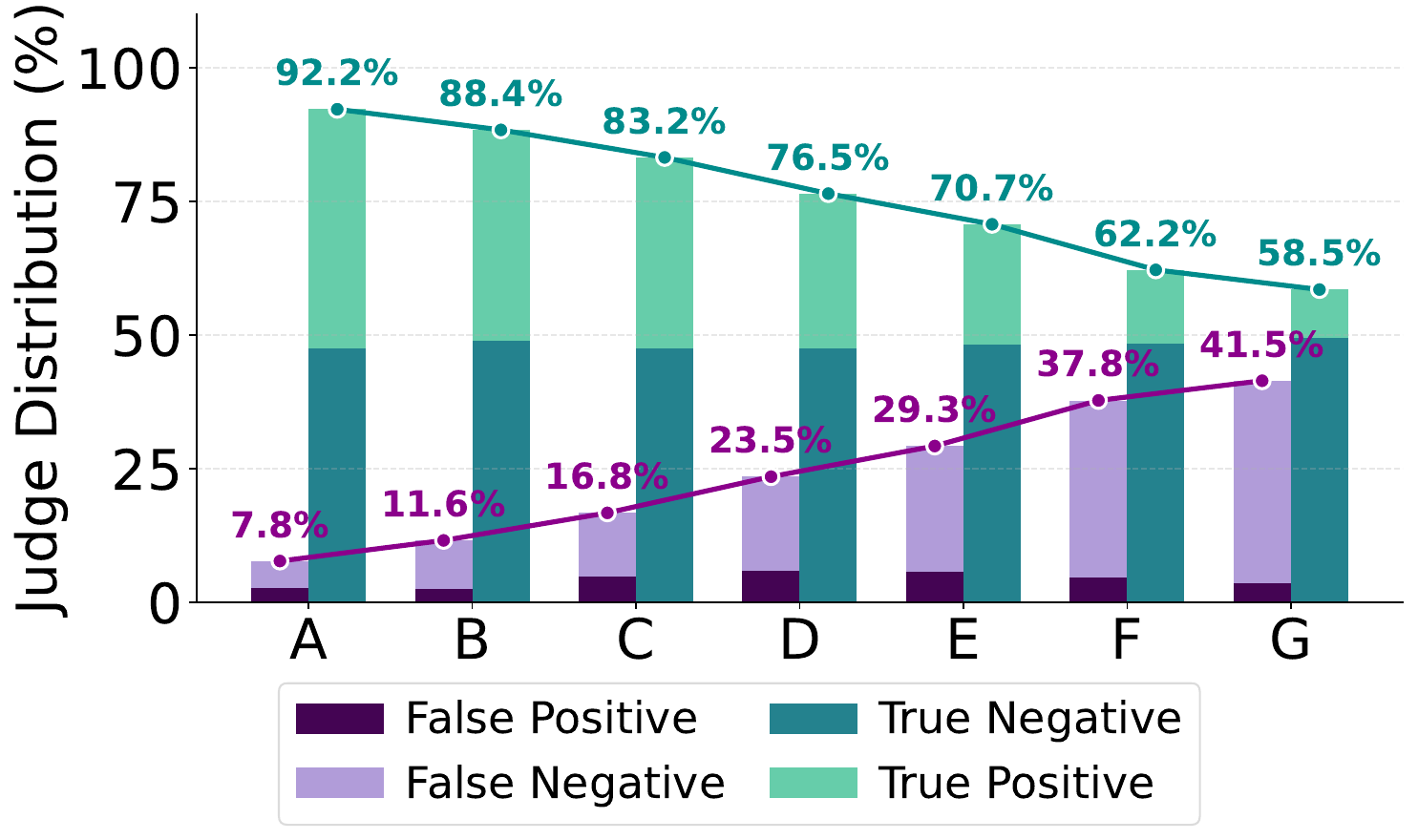}
        \caption{QwQ-32B over human solutions.}
    \end{subfigure}
    \centering
    \begin{subfigure}{0.45\linewidth}
        \includegraphics[width=\linewidth]{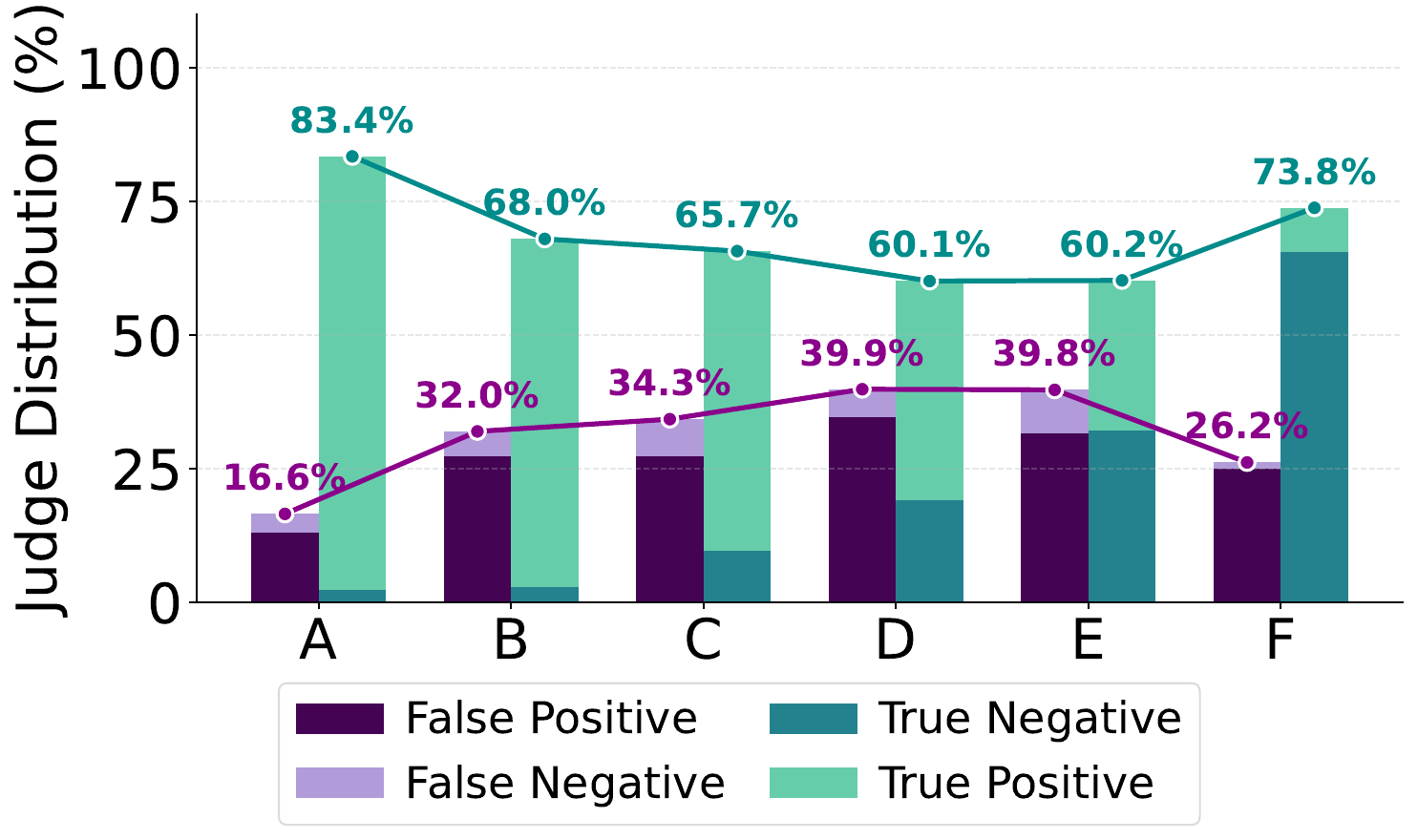}
        \caption{DeepSeek-V3 over self-generated solutions.}
    \end{subfigure}
    \hfill
    \begin{subfigure}{0.45\linewidth}
        \includegraphics[width=\linewidth]{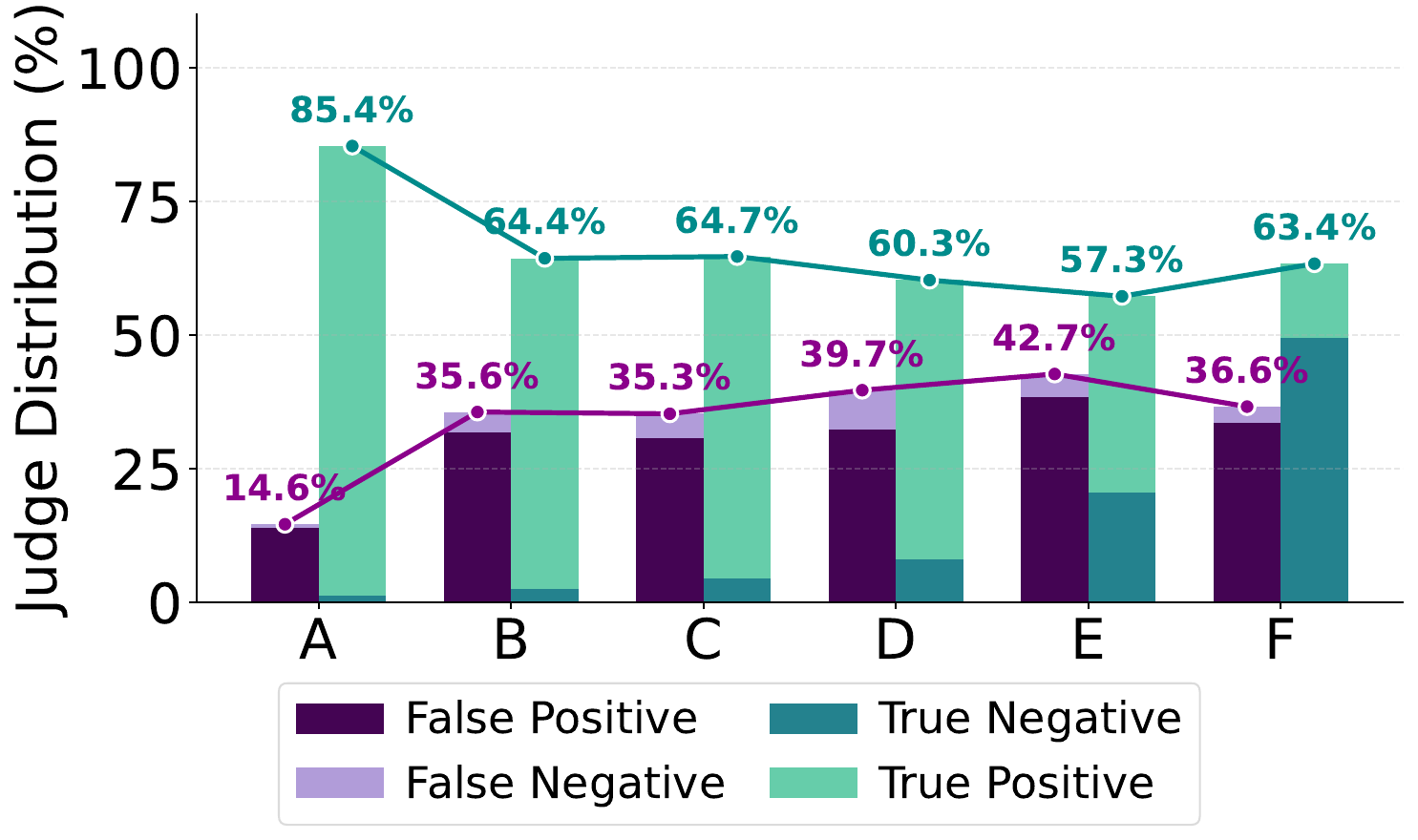}
        \caption{QwQ-32B over self-generated solutions.}
    \end{subfigure}
    \caption{Judge distribution over solutions.}
    \label{fig:cte}
\end{figure*}

\subsection{From Code to X}
\para{\textit{From Code to Editorial: Can the LLM Recognize Its Own Mistakes?}}
\label{sec:cte}

Within our framework, the {editorial} dimension serves as a comprehensive reflection of how understand and analyze the problem. To demonstrate the ability to self-evaluate in LLM, we provide DS-V3 with a candidate code solution and ask it to determine whether the solution is correct. This setup allows us to probe the capacity for error detection in self-generated solutions, thereby shedding light on its potential for self-reflection and self-improvement. We present the results in Figure~\ref{fig:cte}.

When evaluating human-generated solutions, we observe that the judge accuracy consistently decreases as problem difficulty rises. This is expected: as tasks become more complex, the model finds it harder to fully understand the requirements and to follow the logic in human-written solution, much of which may differ from its training data.
However, when the model evaluates its own solutions, we find that its judge accuracy actually increases for the hardest problems. This indicates that, \textbf{it is still able to recognize its own mistakes although the model often produces incorrect code on challenging tasks}. 
This trend is similar to human behavior in that confidence are high for easy problems and very hard ones where failure is expected, and uncertainty peaks for problems of intermediate difficulty whose errors are hard to detect. Interestingly, this kind of self-awareness is also observed in reasoning models like QwQ. The rebound of accuracy in QwQ is less significantly than in DS-V3, highlighting the influence of self-reflection in the reasoning stage. 


\para{\textit{From Code to Cases: Can LLM generate more comprehensive cases with the code?}}

\begin{figure*}[!t]
	\centering
	\begin{subfigure}{0.24\linewidth}
		\includegraphics[width=\linewidth]{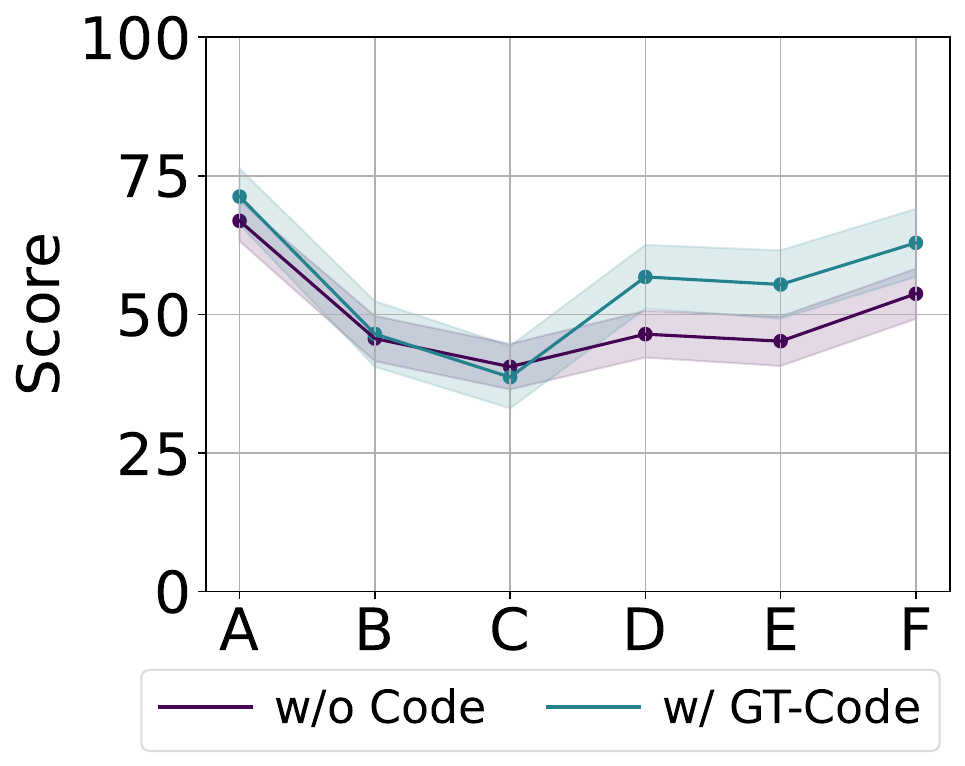}
		\caption{Qwen2.5-72B-Instruct}
	\end{subfigure}
    \hfill
	\begin{subfigure}{0.24\linewidth}
		\includegraphics[width=\linewidth]{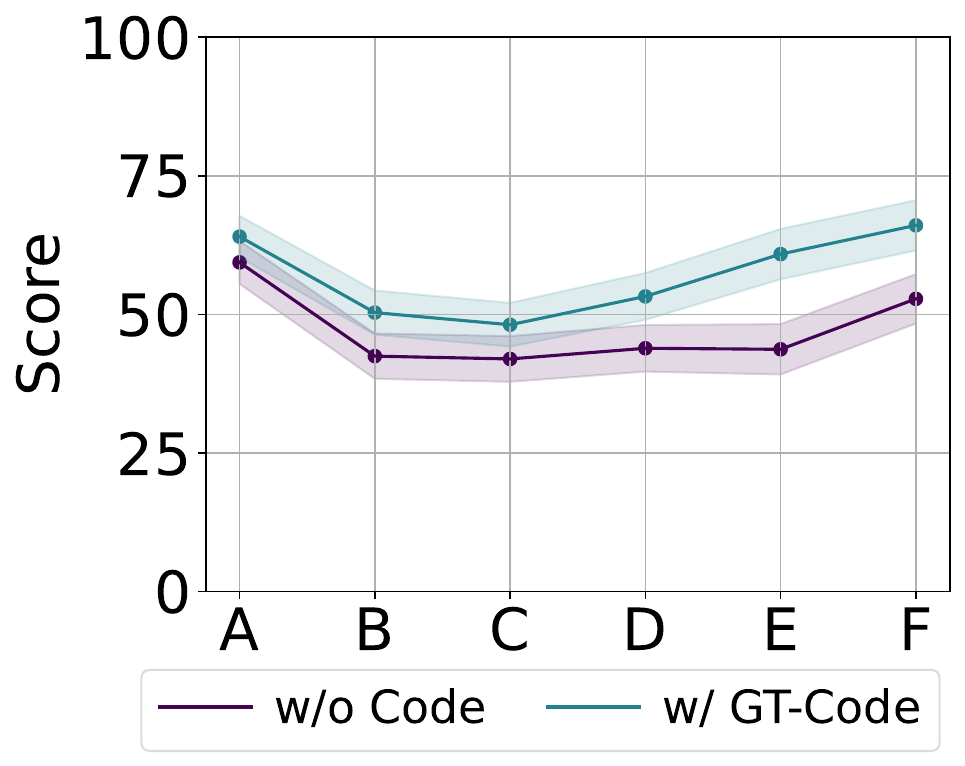}
		\caption{Coder-32B-Instruct}
	\end{subfigure}
    \hfill
    \begin{subfigure}{0.24\linewidth}
        \includegraphics[width=\linewidth]{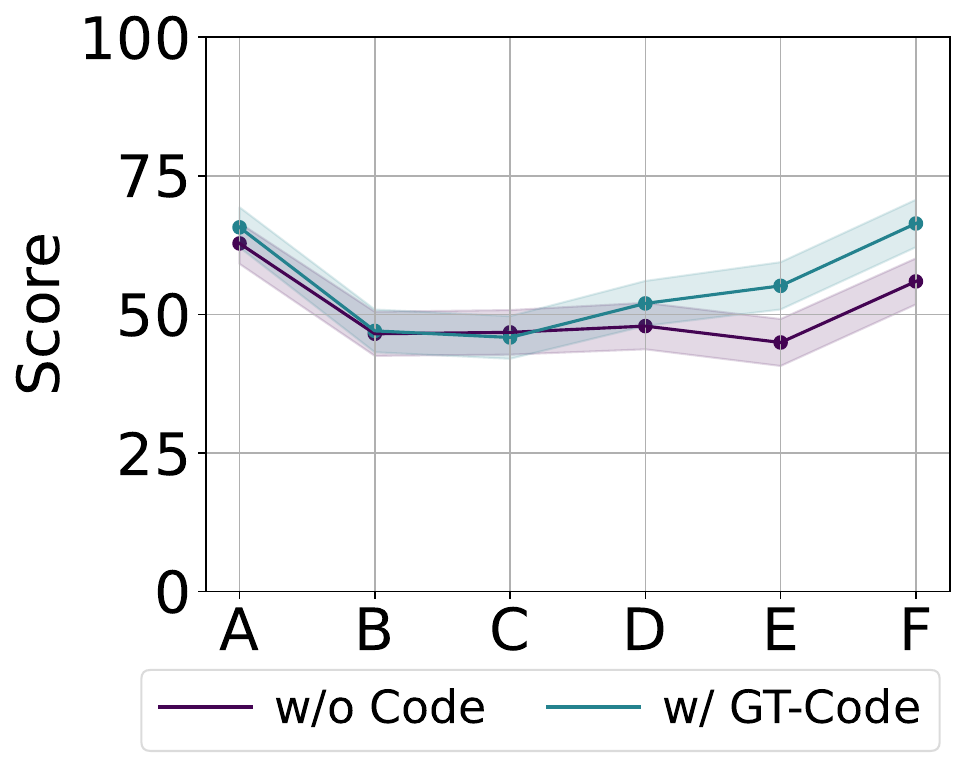}
        \caption{DeepSeek-V3}
    \end{subfigure}
    \hfill
	\begin{subfigure}{0.24\linewidth}
		\includegraphics[width=\linewidth]{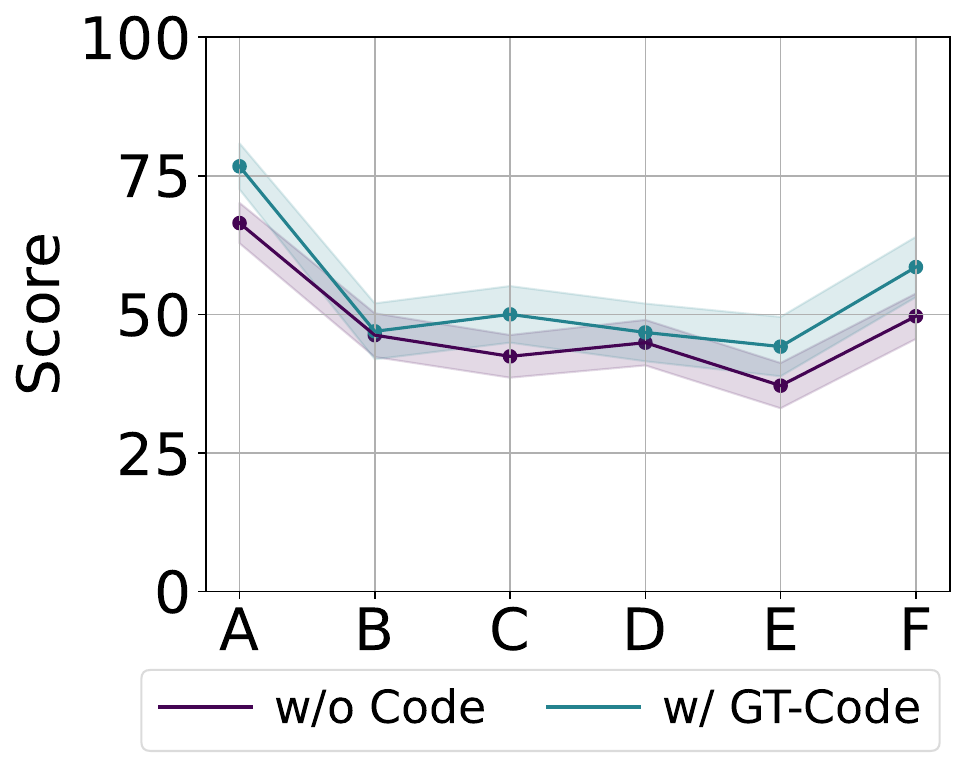}
		\caption{QwQ-32B}
	\end{subfigure}
	\caption{Case Score w.r.t. GT Solutions.}
	\label{fig:all_ctt}
\end{figure*}

Given that the cases is highly related with code, a natural approach is to provide the model with a reference solution. By examining the structure of this solution,  the model can better understand the underlying logic of the inputs and outputs to generate accurate test cases. In our experiments, we prompt the model with correct human-written solutions and present the results in Figure~\ref{fig:all_ctt}.
Our results show that \textbf{providing code solutions significantly improves the accuracy of generated test cases}, particularly for challenging problems like E and F. With access to validated human solutions that handle various edge cases, the model can produce more comprehensive test cases, covering a wide range of scenarios by leveraging this prior knowledge. 
Inspired by these findings, it is potential to generate accurate test cases for problems lacking official cases as offline validation. 
This approach enables richer training data and more robust validation results for code training.

\subsection{From Cases to X}
\para{\textit{From Cases to Editorial: Is LLM also capable to judge cases?}}

\begin{figure*}[!t]
	\centering
    \begin{subfigure}{0.45\linewidth}
        \includegraphics[width=\linewidth]{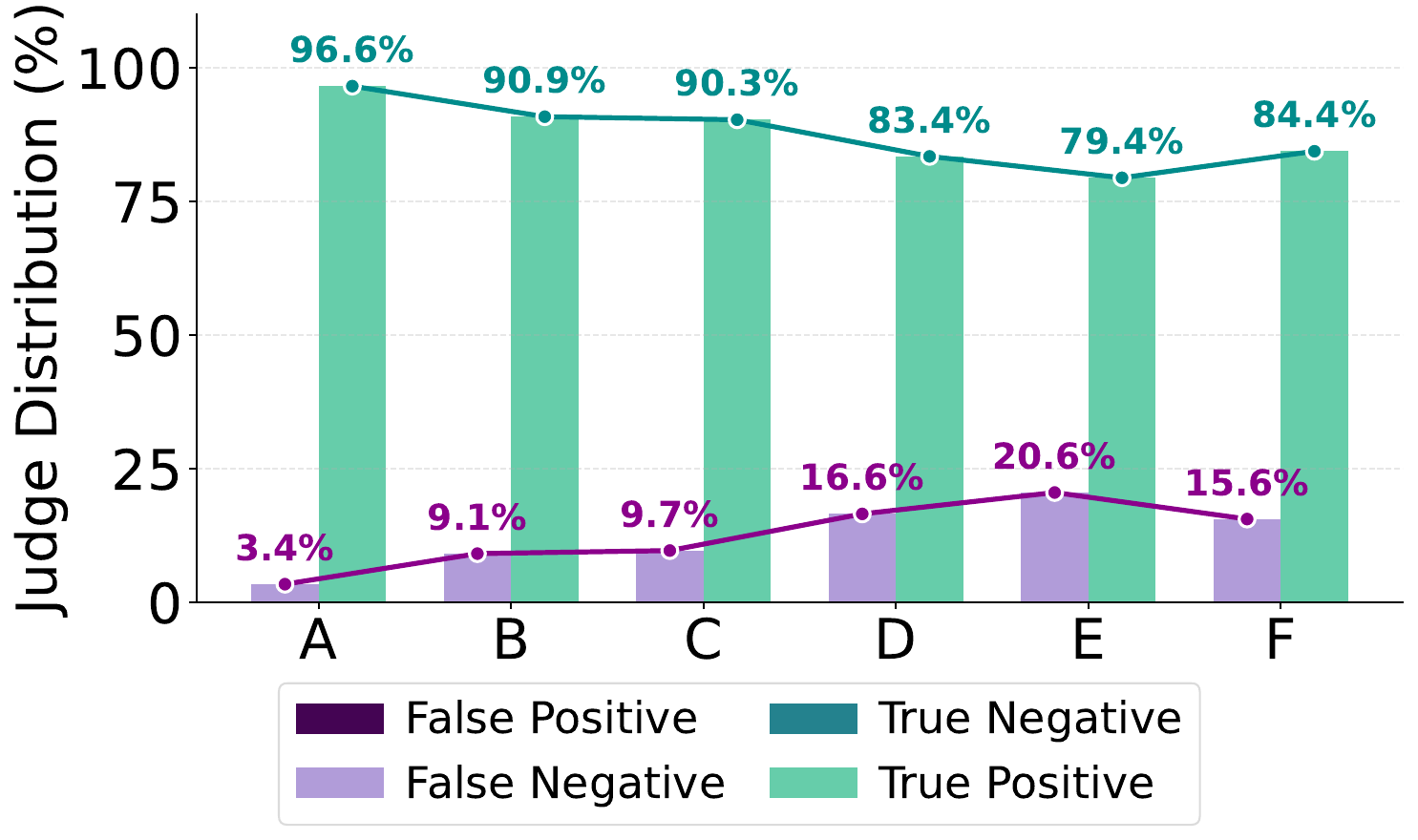}
        \caption{DeepSeek-V3 over ground truth cases.}
    \end{subfigure}
	\hfill
	\begin{subfigure}{0.45\linewidth}
		\includegraphics[width=\linewidth]{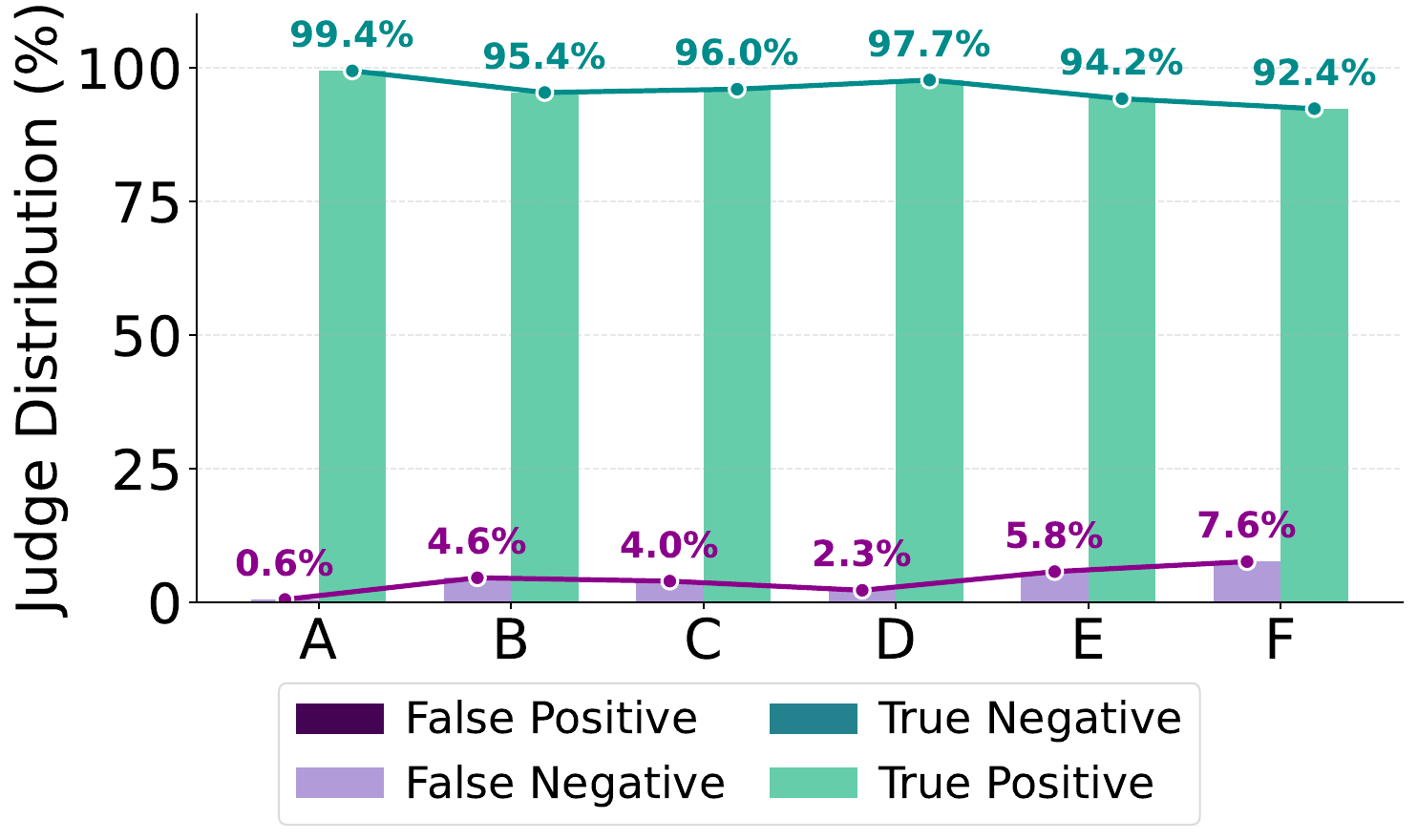}
		\caption{QwQ-32B over ground truth cases.}
	\end{subfigure}

	\centering
    \begin{subfigure}{0.45\linewidth}
        \includegraphics[width=\linewidth]{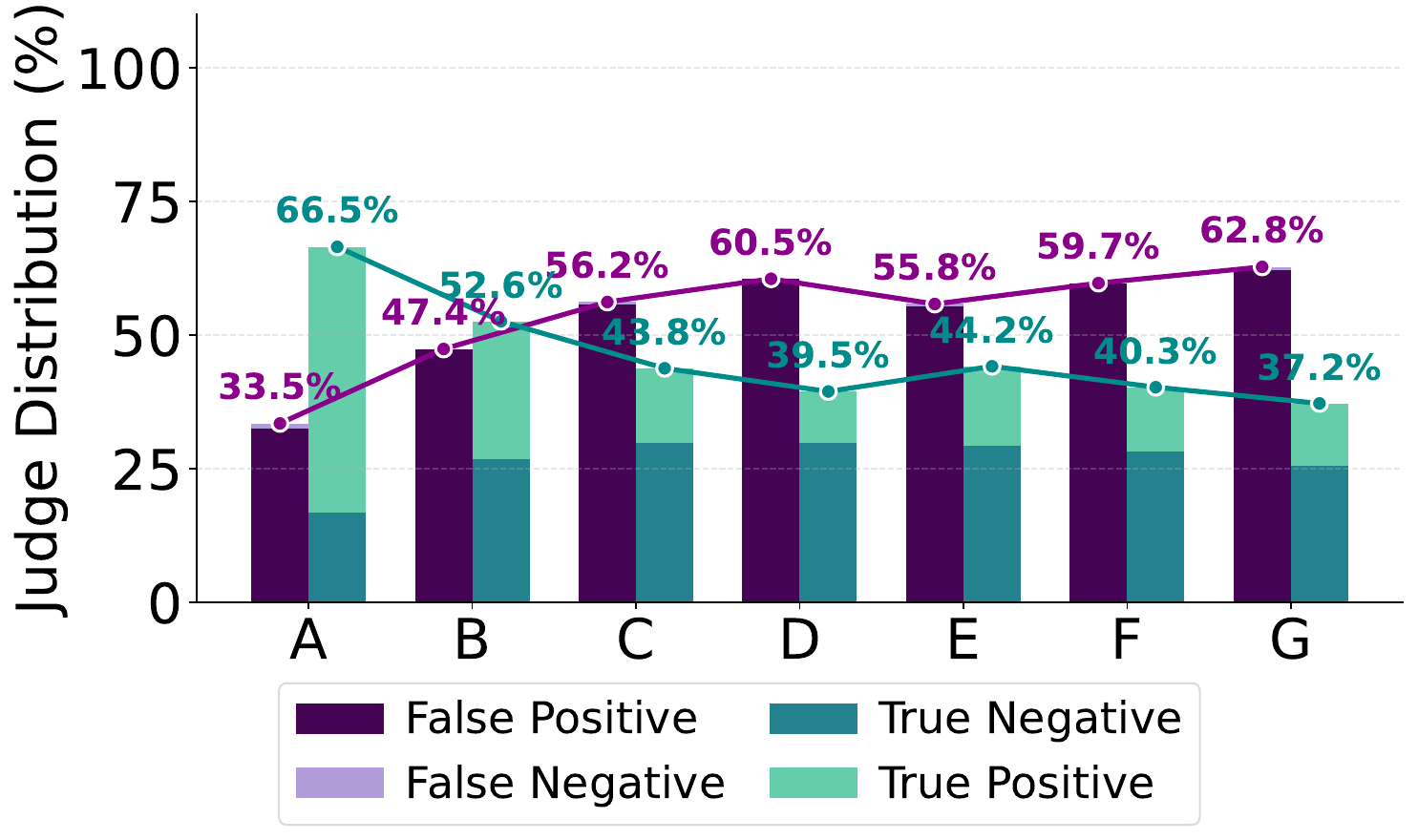}
        \caption{DeepSeek-V3 over self-generated cases.}
    \end{subfigure}
	\hfill
	\begin{subfigure}{0.45\linewidth}
		\includegraphics[width=\linewidth]{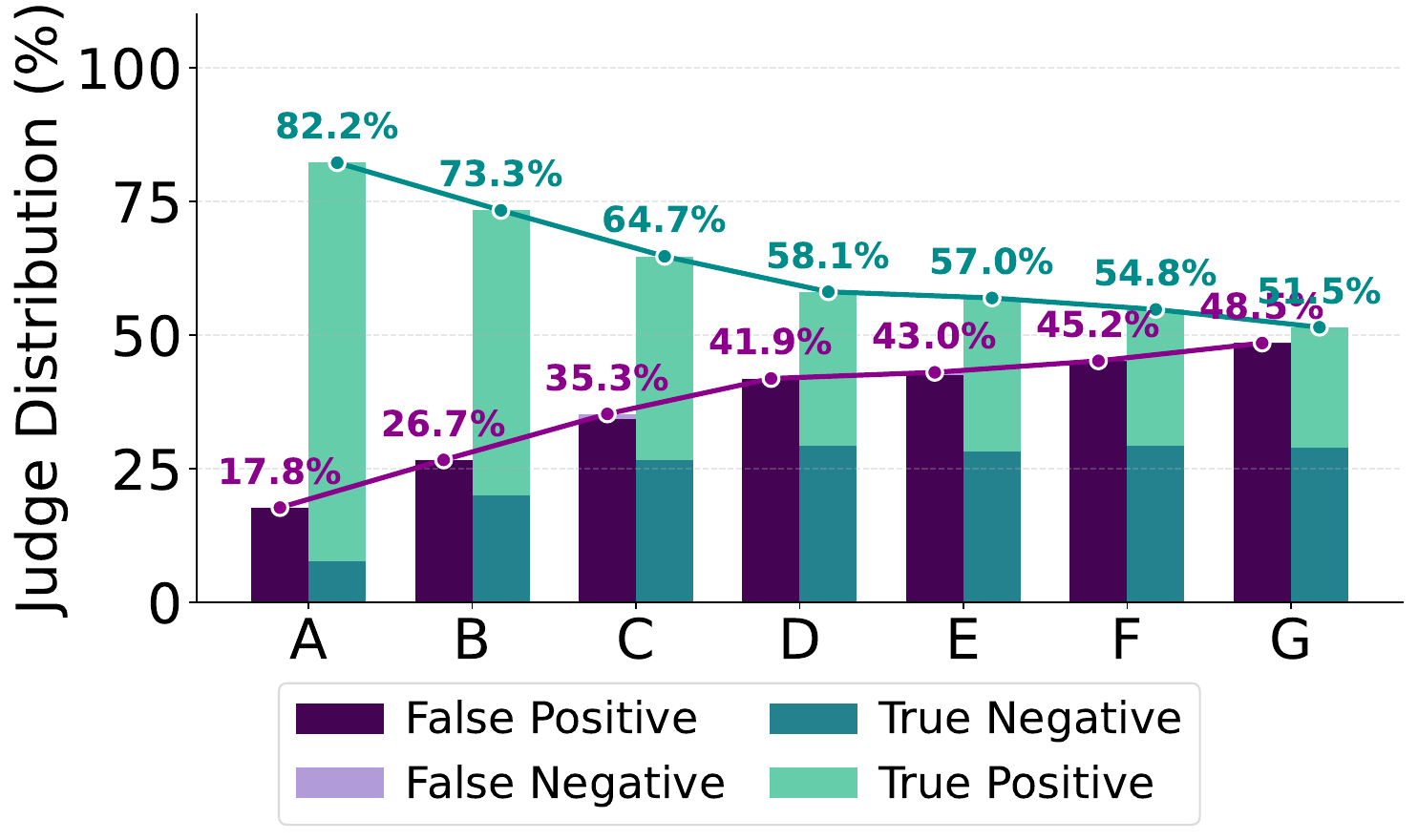}
		\caption{QwQ-32B over self-generated cases.}
	\end{subfigure}
	\caption{Judge accuracy over cases.}
    \label{fig:tte}
\end{figure*}

Following Section~\ref{sec:cte}, we prompt the model to judge the correctness of the test case. The ground truth cases are official and guaranteed to be correct, while the model-generated cases contains both correct and incorrect samples. We exclude test cases whose total string length is larger than 200 and present the results in Figure~\ref{fig:tte}.
We observe that \textbf{the model is able to identify some incorrect cases}. Notably, for ground truth test cases, the model achieves a judging accuracy of up to 90\% for QwQ-32B, and this high accuracy is consistent across different problem difficulties. In contrast, when evaluating its own generated cases, most misjudgments occur when the ground truth case fails but the model incorrectly deems it correct. This suggests that the \textbf{self-consistency still limits its ability to judge cases}, which aligns with the trends observed in Figure~\ref{fig:cte}.

\begin{figure*}[!t]
	\centering
	\begin{subfigure}{0.24\linewidth}
		\includegraphics[width=\linewidth]{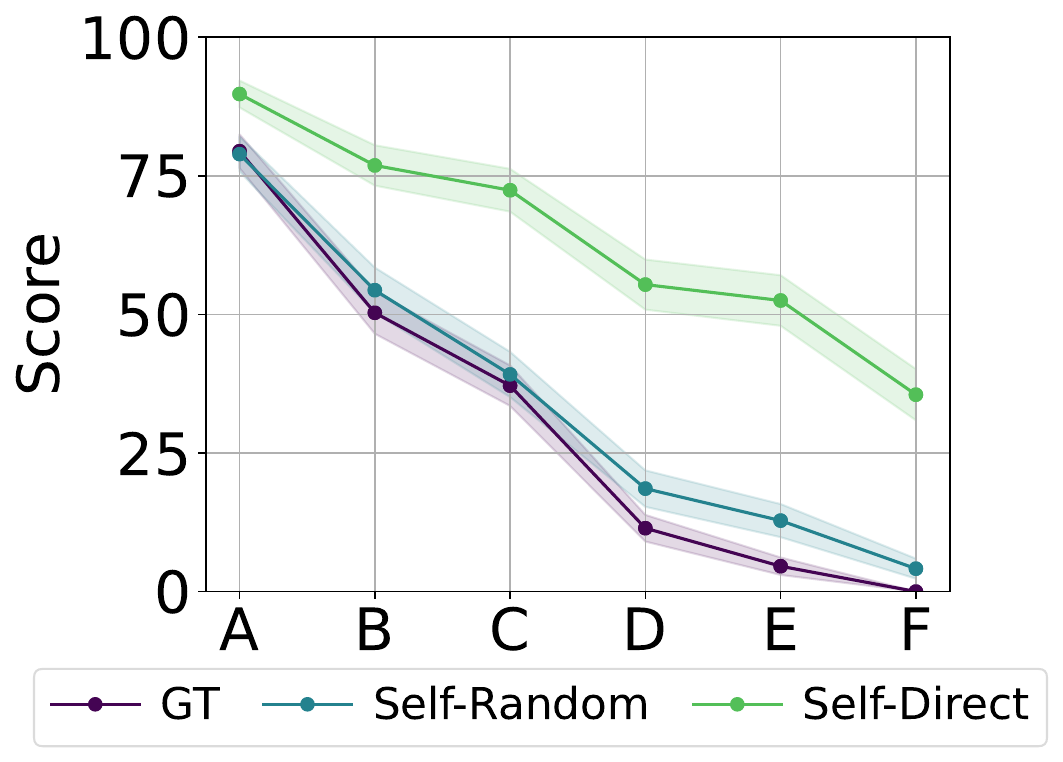}
		\caption{Qwen2.5-72B-Instruct}
		\label{fig:qwen_72b_ttc}
	\end{subfigure}
    \hfill
	\begin{subfigure}{0.24\linewidth}
		\includegraphics[width=\linewidth]{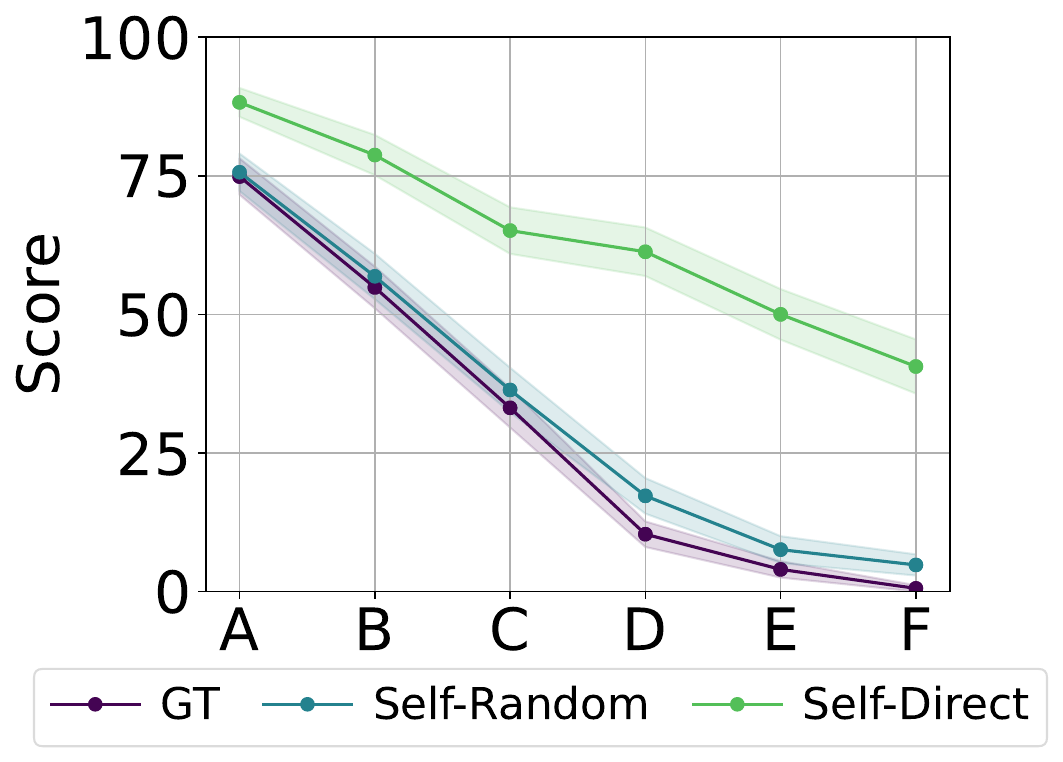}
		\caption{Coder-32B-Instruct}
		\label{fig:coder_32b_ttc}
	\end{subfigure}
    \hfill
    \begin{subfigure}{0.24\linewidth}
        \includegraphics[width=\linewidth]{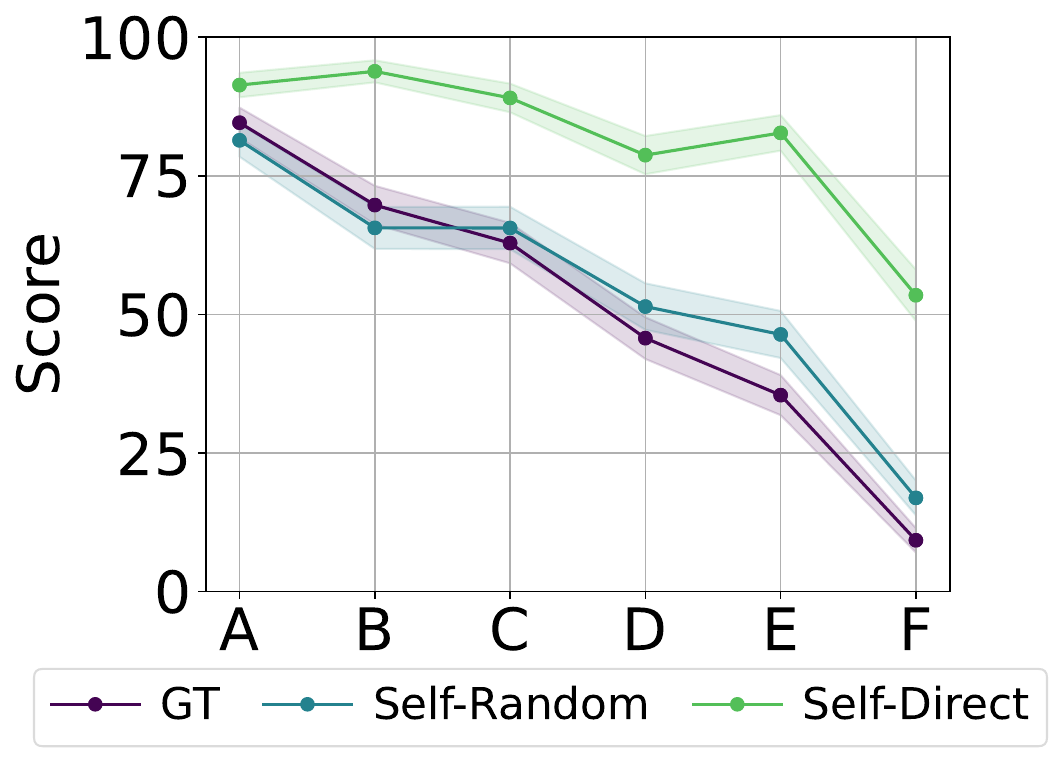}
        \caption{DeepSeek-V3 }
        \label{fig:v3_ttc}
    \end{subfigure}
    \hfill
	\begin{subfigure}{0.24\linewidth}
		\includegraphics[width=\linewidth]{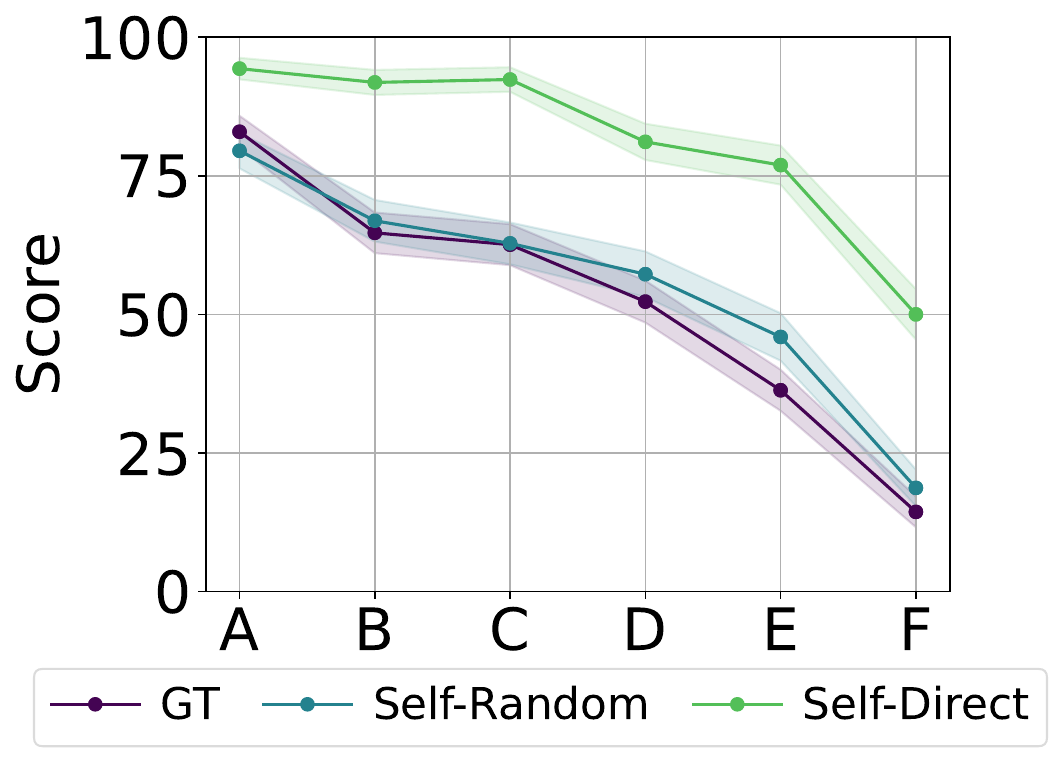}
		\caption{QwQ-32B}
		\label{fig:qwq_32b_ttc}
	\end{subfigure}
	\caption{Pass@1 score over Self-Generated cases.}
	\label{fig:all_ttc}
    \vspace{-0.5em}
\end{figure*}

\para{\textit{From Cases to Code: Can self-generated solution pass all the self-generated cases?}}

In this section, we explore another aspect of self-consistency: whether self-generated solutions can fully pass test cases produced by the model itself.
We create two types of test cases: (1) {Self-Direct}: the model directly generates both inputs and outputs; (2) {Self-Random}: the model generates random inputs, and the outputs are derived from ground truth solutions. 
As shown in Figure~\ref{fig:all_ttc}, \textbf{the pass rate on self-generated test cases is significantly higher than on the ground truth test cases}. This is because self-generated cases often lack comprehensive coverage, particularly of edge conditions, and tend to remain within the bounds of the model's own understanding. Consequently, the accuracy on Self-Random and Self-Direct test cases can exceed that on ground truth cases by up to 5\% and 40\%, respectively.
Interestingly, for some problems that the model fails to solve entirely (e.g., Problem F), it can still achieve an "Accepted" result when evaluated on its own test cases. Nevertheless, the pass rate on self-generated test cases is not 100\% across all problems, indicating that \textbf{these test cases can still uncover errors in the self-generated solutions to some extent}. 
Thus, leveraging self-generated test cases to verify the correctness of solutions remains viable and offers opportunities for self-reflection and iterative self-improvement.

\section{Related Works}
\label{sec:related_work}
\para{Coding Models.} 
Code generation has been fully influenced by LLMs, starting with Codex \cite{chen2021evaluating}, which powers GitHub Copilot and excels in converting natural language prompts into functional code. AlphaCode \cite{li2022competition} leverage an encoder-decoder architecture to solve complex algorithmic problems on platforms like Codeforces. Among open-source models, StarCoder \cite{lozhkov2024starcoder} stands out for its community accessibility and robust performance. Qwen2.5-Coder \cite{hui2024qwen2} achieves impressive results, rivaling GPT-4o, while DeepSeek-R1 \cite{guo2025deepseek} employs reinforcement learning to enhance reasoning and coding capabilities. General-purpose models like GPT-4 \cite{gpt4} and o1 \cite{o1} demonstrate exceptional code generation, highlighting the versatility and rapid evolution of LLMs in this domain.

\para{Coding Evaluation Benchmark.}
Evaluating code generation models relies on robust benchmarks that primarily judge functional correctness. HumanEval \cite{chen2021evaluating} serves as a foundational benchmark with 164 Python problems  measure the ability of generating correct solutions. MBPP \cite{austin2021program} complements HumanEval by providing approximately 1,000 crowd-sourced Python tasks aimed at entry-level programmers. EvalPlus \cite{liu2023your} further enhances evaluation rigor by expanding HumanEval and MBPP with extensive test cases. LiveCodeBench \cite{jain2024livecodebench} introduces a dynamic, contamination-free evaluation with over 880 problems from platforms such as LeetCode and Codeforces, covering diverse tasks including code repair and test output prediction. 
While these benchmarks collectively offer a comprehensive evaluation of code generation capabilities, they mainly focus on evaluating coding solutions and do not consider the model's abilities in editorial analysis or test case generation.
\section{Discussion and Conclusion}
\label{sec:discussion}

\para{Model cognition of LLMs significantly differs from human distribution.} 
Our large-scale experiments reveal that the distribution of solutions and cases they produce significantly differs from that of real-world data from human. 
This phenomenon indicates that the internal cognitive framework shaped by training data imposes constraints, limiting the ability to develop creative reasoning pathways and invent novel approaches when solving the code challenge.

\para{Self-consistency exists within model cognition.}
Analyzing the prediction across different dimensions, we observe that self-consistency exists within the model cognition. For example, leveraging the self-generated editorial does not lead to significant improvements in solutions. Besides, the roll-out solutions can easily pass the self-generated test cases, resulting in a kind of reward hacking with respect to test cases. These observations indicate that, LLM exhibits self-consistency within its cognitive stage due to limitations of its own training data.

\para{Inconsistency across various dimensions may facilitate self-improvement.} 
LLM demonstrates different performance across the dimensions of the Coding Triangle and these differences can be further harnessed for self-improvement, such as using its own judgment to distinguish correct solutions, or leveraging a correct code to improve edge-case generation.
This approach indicates that the development of self-improvement can be realized through iteratively align these dimensions, gradually reducing error correlations and develop LLMs with more powerful coding ability.

\para{Model mixture enhance diversity and robustness.} 
Recognizing the existence of self-consistency, we find that the combination of different models can significantly reduce bias while improving both robustness and diversity. For example, solutions generated by different models can help identify a wider range of boundary cases, and the combined test cases from multiple models are effective at detecting potential errors. These results demonstrate that leveraging model mixtures mitigates the biases and diverse outputs from different models contribute to better robustness.

\para{Limitation and Conclusion.}
We systematically investigate the coding ability of LLMs through the lens of the coding triangle, i.e., editorial, code, and cases. 
There are still some limitations as we does not fully explore all possible interactions between these three dimensions.
Overall, our study reveal the self-consistency and self-inconsistency inside LLMs, point out the importance of bridging the gap between model cognition and human expertise, and provide a potential direction of aligning and mutually reinforcing the three dimensions to achieve more reliable and generalizable coding models.

\bibliographystyle{plainnat}
\bibliography{main}

\begin{thebibliography}{32}
\providecommand{\natexlab}[1]{#1}
\providecommand{\url}[1]{\texttt{#1}}
\expandafter\ifx\csname urlstyle\endcsname\relax
  \providecommand{\doi}[1]{doi: #1}\else
  \providecommand{\doi}{doi: \begingroup \urlstyle{rm}\Url}\fi

\bibitem[Achiam et~al.(2023)Achiam, Adler, Agarwal, Ahmad, Akkaya, Aleman, Almeida, Altenschmidt, Altman, Anadkat, et~al.]{gpt4}
Josh Achiam, Steven Adler, Sandhini Agarwal, Lama Ahmad, Ilge Akkaya, Florencia~Leoni Aleman, Diogo Almeida, Janko Altenschmidt, Sam Altman, Shyamal Anadkat, et~al.
\newblock Gpt-4 technical report.
\newblock \emph{arXiv preprint arXiv:2303.08774}, 2023.

\bibitem[Anthropic(2024)]{claude3.5}
Anthropic.
\newblock Claude 3.5 sonnet.
\newblock \url{https://www.anthropic.com/news/claude-3-5-sonnet}, 2024.
\newblock 2024.06.21.

\bibitem[Austin et~al.(2021{\natexlab{a}})Austin, Odena, Nye, Bosma, Michalewski, Dohan, Jiang, Cai, Terry, Le, et~al.]{austin2021program}
Jacob Austin, Augustus Odena, Maxwell Nye, Maarten Bosma, Henryk Michalewski, David Dohan, Ellen Jiang, Carrie Cai, Michael Terry, Quoc Le, et~al.
\newblock Program synthesis with large language models.
\newblock \emph{arXiv preprint arXiv:2108.07732}, 2021{\natexlab{a}}.

\bibitem[Austin et~al.(2021{\natexlab{b}})Austin, Odena, Nye, Bosma, Michalewski, Dohan, Jiang, Cai, Terry, Le, et~al.]{austin2021programsynthesislargelanguage}
Jacob Austin, Augustus Odena, Maxwell Nye, Maarten Bosma, Henryk Michalewski, David Dohan, Ellen Jiang, Carrie Cai, Michael Terry, Quoc Le, et~al.
\newblock Program synthesis with large language models.
\newblock \emph{arXiv preprint arXiv:2108.07732}, 2021{\natexlab{b}}.

\bibitem[Bai et~al.(2023)Bai, Bai, Chu, Cui, Dang, Deng, Fan, Ge, Han, Huang, et~al.]{qwen}
Jinze Bai, Shuai Bai, Yunfei Chu, Zeyu Cui, Kai Dang, Xiaodong Deng, Yang Fan, Wenbin Ge, Yu~Han, Fei Huang, et~al.
\newblock Qwen technical report.
\newblock \emph{arXiv preprint arXiv:2309.16609}, 2023.

\bibitem[Brown(2020)]{gpt3}
Tom~B Brown.
\newblock Language models are few-shot learners.
\newblock \emph{arXiv preprint arXiv:2005.14165}, 2020.

\bibitem[Cassano et~al.(2022)Cassano, Gouwar, Nguyen, Nguyen, Phipps-Costin, Pinckney, Yee, Zi, Anderson, Feldman, et~al.]{cassano2022multiplescalableextensibleapproach}
Federico Cassano, John Gouwar, Daniel Nguyen, Sydney Nguyen, Luna Phipps-Costin, Donald Pinckney, Ming-Ho Yee, Yangtian Zi, Carolyn~Jane Anderson, Molly~Q Feldman, et~al.
\newblock Multipl-e: A scalable and extensible approach to benchmarking neural code generation.
\newblock \emph{arXiv preprint arXiv:2208.08227}, 2022.

\bibitem[Chen et~al.(2021{\natexlab{a}})Chen, Tworek, Jun, Yuan, Pinto, Kaplan, Edwards, Burda, Joseph, Brockman, et~al.]{chen2021evaluating}
Mark Chen, Jerry Tworek, Heewoo Jun, Qiming Yuan, Henrique Ponde De~Oliveira Pinto, Jared Kaplan, Harri Edwards, Yuri Burda, Nicholas Joseph, Greg Brockman, et~al.
\newblock Evaluating large language models trained on code.
\newblock \emph{arXiv preprint arXiv:2107.03374}, 2021{\natexlab{a}}.

\bibitem[Chen et~al.(2021{\natexlab{b}})Chen, Tworek, Jun, Yuan, Pinto, Kaplan, Edwards, Burda, Joseph, Brockman, et~al.]{chen2021evaluatinglargelanguagemodels}
Mark Chen, Jerry Tworek, Heewoo Jun, Qiming Yuan, Henrique Ponde De~Oliveira Pinto, Jared Kaplan, Harri Edwards, Yuri Burda, Nicholas Joseph, Greg Brockman, et~al.
\newblock Evaluating large language models trained on code.
\newblock \emph{arXiv preprint arXiv:2107.03374}, 2021{\natexlab{b}}.

\bibitem[Dubey et~al.(2024)Dubey, Jauhri, Pandey, Kadian, Al-Dahle, Letman, Mathur, Schelten, Yang, Fan, et~al.]{llama3}
Abhimanyu Dubey, Abhinav Jauhri, Abhinav Pandey, Abhishek Kadian, Ahmad Al-Dahle, Aiesha Letman, Akhil Mathur, Alan Schelten, Amy Yang, Angela Fan, et~al.
\newblock The llama 3 herd of models.
\newblock \emph{arXiv preprint arXiv:2407.21783}, 2024.

\bibitem[El-Kishky et~al.(2025)El-Kishky, Wei, Saraiva, Minaiev, Selsam, Dohan, Song, Lightman, Clavera, Pachocki, et~al.]{el2025competitive}
Ahmed El-Kishky, Alexander Wei, Andre Saraiva, Borys Minaiev, Daniel Selsam, David Dohan, Francis Song, Hunter Lightman, Ignasi Clavera, Jakub Pachocki, et~al.
\newblock Competitive programming with large reasoning models.
\newblock \emph{arXiv preprint arXiv:2502.06807}, 2025.

\bibitem[Guo et~al.(2025)Guo, Yang, Zhang, Song, Zhang, Xu, Zhu, Ma, Wang, Bi, et~al.]{guo2025deepseek}
Daya Guo, Dejian Yang, Haowei Zhang, Junxiao Song, Ruoyu Zhang, Runxin Xu, Qihao Zhu, Shirong Ma, Peiyi Wang, Xiao Bi, et~al.
\newblock Deepseek-r1: Incentivizing reasoning capability in llms via reinforcement learning.
\newblock \emph{arXiv preprint arXiv:2501.12948}, 2025.

\bibitem[Guo et~al.(2024)Guo, Li, Liu, Ma, Zheng, Yu, Pan, Li, Liu, Wang, et~al.]{codeeditorbench}
Jiawei Guo, Ziming Li, Xueling Liu, Kaijing Ma, Tianyu Zheng, Zhouliang Yu, Ding Pan, Yizhi Li, Ruibo Liu, Yue Wang, et~al.
\newblock Codeeditorbench: Evaluating code editing capability of large language models.
\newblock \emph{arXiv preprint arXiv:2404.03543}, 2024.

\bibitem[Hui et~al.(2024)Hui, Yang, Cui, Yang, Liu, Zhang, Liu, Zhang, Yu, Lu, et~al.]{hui2024qwen2}
Binyuan Hui, Jian Yang, Zeyu Cui, Jiaxi Yang, Dayiheng Liu, Lei Zhang, Tianyu Liu, Jiajun Zhang, Bowen Yu, Keming Lu, et~al.
\newblock Qwen2. 5-coder technical report.
\newblock \emph{arXiv preprint arXiv:2409.12186}, 2024.

\bibitem[Jain et~al.(2024)Jain, Han, Gu, Li, Yan, Zhang, Wang, Solar-Lezama, Sen, and Stoica]{jain2024livecodebench}
Naman Jain, King Han, Alex Gu, Wen-Ding Li, Fanjia Yan, Tianjun Zhang, Sida Wang, Armando Solar-Lezama, Koushik Sen, and Ion Stoica.
\newblock Livecodebench: Holistic and contamination free evaluation of large language models for code.
\newblock \emph{arXiv preprint arXiv:2403.07974}, 2024.

\bibitem[Jiang et~al.(2023)Jiang, Sablayrolles, Mensch, Bamford, Chaplot, de~las Casas, Bressand, Lengyel, Lample, Saulnier, et~al.]{mistral}
AQ~Jiang, A~Sablayrolles, A~Mensch, C~Bamford, DS~Chaplot, D~de~las Casas, F~Bressand, G~Lengyel, G~Lample, L~Saulnier, et~al.
\newblock Mistral 7b (2023).
\newblock \emph{arXiv preprint arXiv:2310.06825}, 2023.

\bibitem[Li et~al.(2022)Li, Choi, Chung, Kushman, Schrittwieser, Leblond, Eccles, Keeling, Gimeno, Dal~Lago, et~al.]{li2022competition}
Yujia Li, David Choi, Junyoung Chung, Nate Kushman, Julian Schrittwieser, R{\'e}mi Leblond, Tom Eccles, James Keeling, Felix Gimeno, Agustin Dal~Lago, et~al.
\newblock Competition-level code generation with alphacode.
\newblock \emph{Science}, 378\penalty0 (6624):\penalty0 1092--1097, 2022.

\bibitem[Li et~al.(2024)Li, Zang, Ma, Guo, Zheng, Niu, Yue, Wang, Yang, Liu, et~al.]{autokaggle}
Ziming Li, Qianbo Zang, David Ma, Jiawei Guo, Tianyu Zheng, Xinyao Niu, Xiang Yue, Yue Wang, Jian Yang, Jiaheng Liu, et~al.
\newblock Autokaggle: A multi-agent framework for autonomous data science competitions.
\newblock \emph{arXiv preprint arXiv:2410.20424}, 2024.

\bibitem[Liu et~al.(2024{\natexlab{a}})Liu, Feng, Xue, Wang, Wu, Lu, Zhao, Deng, Zhang, Ruan, et~al.]{liu2024deepseek}
Aixin Liu, Bei Feng, Bing Xue, Bingxuan Wang, Bochao Wu, Chengda Lu, Chenggang Zhao, Chengqi Deng, Chenyu Zhang, Chong Ruan, et~al.
\newblock Deepseek-v3 technical report.
\newblock \emph{arXiv preprint arXiv:2412.19437}, 2024{\natexlab{a}}.

\bibitem[Liu et~al.(2024{\natexlab{b}})Liu, Deng, Liu, Yang, Liu, Zhu, Zhao, Chai, Wu, Jin, et~al.]{m2rceval}
Jiaheng Liu, Ken Deng, Congnan Liu, Jian Yang, Shukai Liu, He~Zhu, Peng Zhao, Linzheng Chai, Yanan Wu, Ke~Jin, et~al.
\newblock M2rc-eval: Massively multilingual repository-level code completion evaluation.
\newblock \emph{arXiv preprint arXiv:2410.21157}, 2024{\natexlab{b}}.

\bibitem[Liu et~al.(2023)Liu, Xia, Wang, and Zhang]{liu2023your}
Jiawei Liu, Chunqiu~Steven Xia, Yuyao Wang, and Lingming Zhang.
\newblock Is your code generated by chatgpt really correct? rigorous evaluation of large language models for code generation.
\newblock \emph{Advances in Neural Information Processing Systems}, 36:\penalty0 21558--21572, 2023.

\bibitem[Lozhkov et~al.(2024)Lozhkov, Li, Allal, Cassano, Lamy-Poirier, Tazi, Tang, Pykhtar, Liu, Wei, et~al.]{lozhkov2024starcoder}
Anton Lozhkov, Raymond Li, Loubna~Ben Allal, Federico Cassano, Joel Lamy-Poirier, Nouamane Tazi, Ao~Tang, Dmytro Pykhtar, Jiawei Liu, Yuxiang Wei, et~al.
\newblock Starcoder 2 and the stack v2: The next generation.
\newblock \emph{arXiv preprint arXiv:2402.19173}, 2024.

\bibitem[OpenAI(2024)]{gpt4o}
OpenAI.
\newblock Gpt-4o.
\newblock \url{https://openai.com/index/hello-gpt-4o}, 2024.
\newblock 2024.05.13.

\bibitem[OpenAI(2025{\natexlab{a}})]{o1}
OpenAI.
\newblock Openai o1 system card.
\newblock \url{https://openai.com/index/openai-o1-system-card/}, 2025{\natexlab{a}}.

\bibitem[OpenAI(2025{\natexlab{b}})]{o3}
OpenAI.
\newblock Openai o3 system card.
\newblock \url{https://openai.com/index/o3-o4-mini-system-card/}, 2025{\natexlab{b}}.

\bibitem[OpenAI(2025{\natexlab{c}})]{o3mini}
OpenAI.
\newblock Openai o3-mini system card.
\newblock \url{https://openai.com/index/openai-o3-mini/}, 2025{\natexlab{c}}.

\bibitem[Team(2025{\natexlab{a}})]{qwen3}
Qwen Team.
\newblock Qwen3: Think deeper, act faster.
\newblock \url{https://qwenlm.github.io/blog/qwen3/}, 2025{\natexlab{a}}.

\bibitem[Team(2025{\natexlab{b}})]{qwq32b}
Qwen Team.
\newblock Qwq-32b: Embracing the power of reinforcement learning, March 2025{\natexlab{b}}.
\newblock URL \url{https://qwenlm.github.io/blog/qwq-32b/}.

\bibitem[Touvron et~al.(2023)Touvron, Martin, Stone, Albert, Almahairi, Babaei, Bashlykov, Batra, Bhargava, Bhosale, et~al.]{llama2}
Hugo Touvron, Louis Martin, Kevin Stone, Peter Albert, Amjad Almahairi, Yasmine Babaei, Nikolay Bashlykov, Soumya Batra, Prajjwal Bhargava, Shruti Bhosale, et~al.
\newblock Llama 2: Open foundation and fine-tuned chat models.
\newblock \emph{arXiv preprint arXiv:2307.09288}, 2023.

\bibitem[Wu et~al.(2024)Wu, Yang, Chai, Zhang, Liu, Du, Liang, Shu, Cheng, Sun, et~al.]{tablebench}
Xianjie Wu, Jian Yang, Linzheng Chai, Ge~Zhang, Jiaheng Liu, Xinrun Du, Di~Liang, Daixin Shu, Xianfu Cheng, Tianzhen Sun, et~al.
\newblock Tablebench: A comprehensive and complex benchmark for table question answering.
\newblock \emph{arXiv preprint arXiv:2408.09174}, 2024.

\bibitem[Yang et~al.(2024{\natexlab{a}})Yang, Yang, Hui, Zheng, Yu, Zhou, Li, Li, Liu, Huang, et~al.]{qwen2}
An~Yang, Baosong Yang, Binyuan Hui, Bo~Zheng, Bowen Yu, Chang Zhou, Chengpeng Li, Chengyuan Li, Dayiheng Liu, Fei Huang, et~al.
\newblock Qwen2 technical report.
\newblock \emph{arXiv preprint arXiv:2407.10671}, 2024{\natexlab{a}}.

\bibitem[Yang et~al.(2024{\natexlab{b}})Yang, Yang, Zhang, Hui, Zheng, Yu, Li, Liu, Huang, Wei, et~al.]{yang2024qwen2}
An~Yang, Baosong Yang, Beichen Zhang, Binyuan Hui, Bo~Zheng, Bowen Yu, Chengyuan Li, Dayiheng Liu, Fei Huang, Haoran Wei, et~al.
\newblock Qwen2. 5 technical report.
\newblock \emph{arXiv preprint arXiv:2412.15115}, 2024{\natexlab{b}}.

\end{thebibliography}

\clearpage

\appendix
\section*{\Large{Appendix}}
\label{sec:Appendix}
\section{Dataset and License}

The evaluating problem dataset is collected from AtCoder (\url{https://atcoder.jp}).
We only collect publicly available content, including visible editorials, sample codes, and test cases published on the website.
All collected materials are strictly used for research and evaluation purposes only, specifically to assess the performance of candidate models.
No part of the collected dataset is used for training any model and we fully respect the content ownership and terms of use of AtCoder.
We follow LiveCodeBench and abide by Fair Use §107: "the fair use of a copyrighted work, including such use by ... scholarship, or research, is not an infringement of copyright", where fair use is
determined by "the purpose and character of the use, including whether such use is of a commercial
nature or is for nonprofit educational purposes" and "the effect of the use upon the potential market
for or value of the copyrighted work."

\section{More Experiments}
In this section, we provide more experiments as additional results. We mainly focus on model mixture and the interaction across different models. 



\subsection{Does LLM benefit from the editorial from other models?}

\begin{figure*}[!h]
	\centering
	\begin{subfigure}{0.3\linewidth}
		\includegraphics[width=\linewidth]{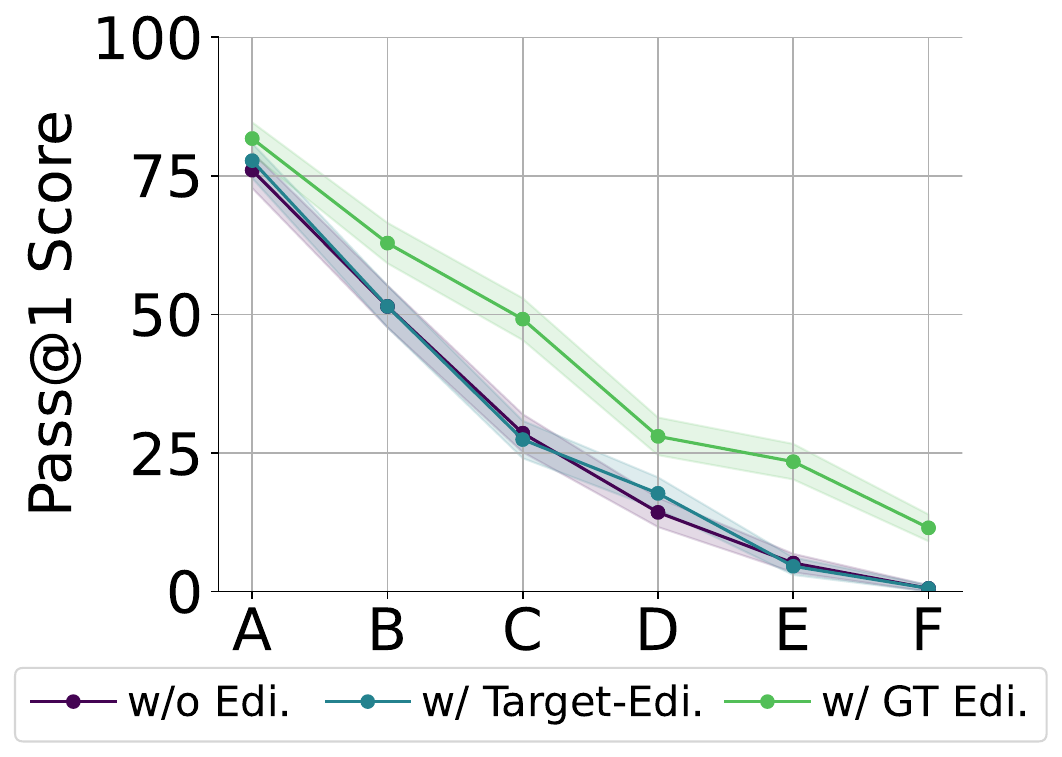}
		\caption{Qwen2.5-72B-Instruct}
		\label{fig:qwen_72b_etc_cross}
	\end{subfigure}
    \hfill
	\begin{subfigure}{0.3\linewidth}
		\includegraphics[width=\linewidth]{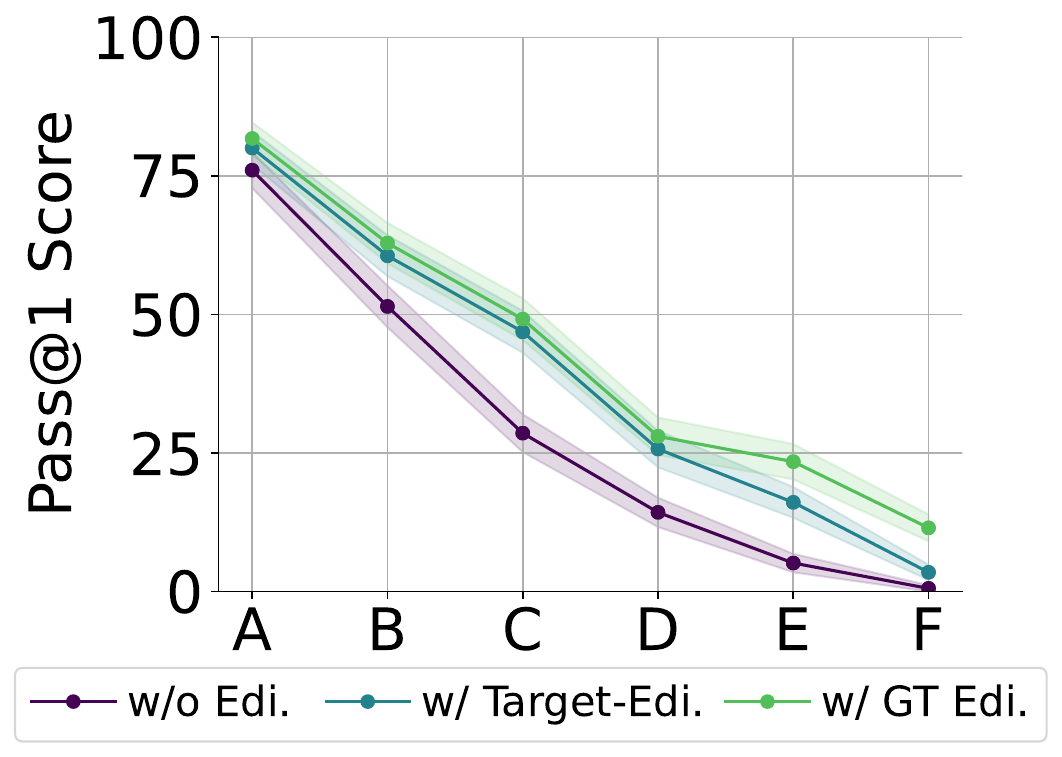}
		\caption{DeepSeek-V3}
		\label{fig:ds_v3_etc_cross}
	\end{subfigure}
    \hfill
    \begin{subfigure}{0.3\linewidth}
        \includegraphics[width=\linewidth]{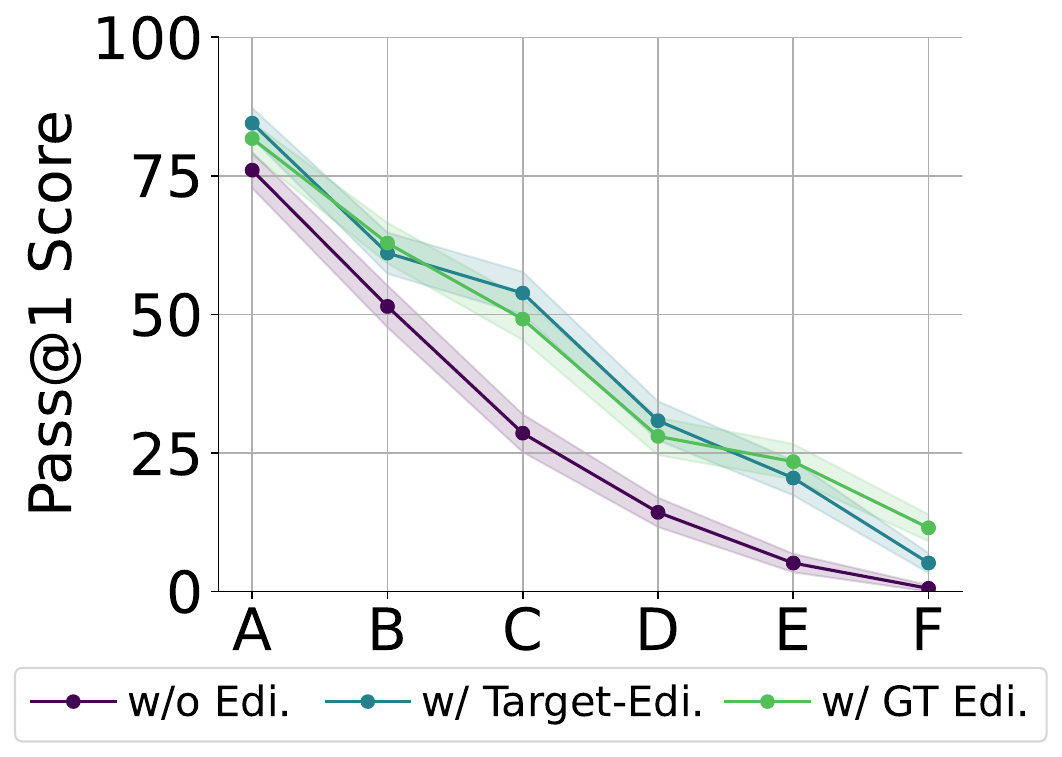}
        \caption{QwQ-32B}
        \label{fig:qwq_32b_etc_cross}
    \end{subfigure}
	\caption{Pass@1 score of Coder-32B-Instruct with editorials from other models.}
	\label{fig:etc_cross}
\end{figure*}

We further investigate whether editorials generated by other models can benefit code generation. To this end, we evaluate the performance of Coder-32B-Instruct when provided with editorials from Qwen2.5-72B-Instruct, DeepSeek-V3, and QwQ-32B, as shown in Figure~\ref{fig:etc_cross}. The results indicate that using editorials as prompts can serve as a form of knowledge distillation, enhancing the performance of the student model. With access to relatively accurate editorials from reasoning-oriented models, Coder-32B-Instruct, which excels at code implementation, can achieve performance comparable to that obtained with ground truth editorials, demonstrating the practicality of leveraging logical knowledge embedded in reasoning models to improve the code generation capabilities of coding models.

\subsection{Can LLM recognize mistakes from other models?}
\begin{figure*}[!h]
	\centering
	\begin{subfigure}{0.49\linewidth}
		\includegraphics[width=\linewidth]{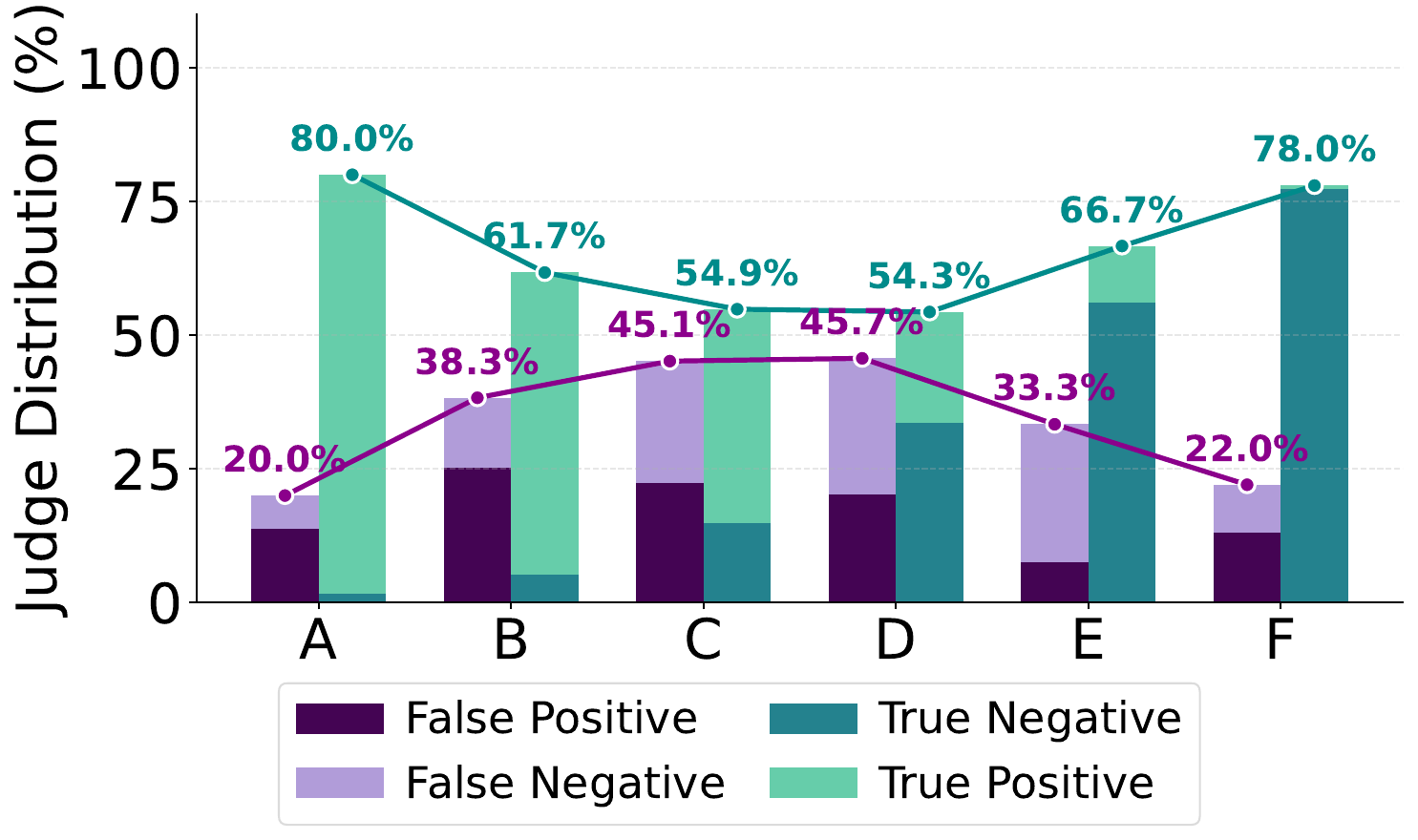}
		\caption{Qwen2.5-72B-Instruct}
		\label{fig:qwen_72b_etc_cross}
	\end{subfigure}
    \hfill
	\begin{subfigure}{0.49\linewidth}
		\includegraphics[width=\linewidth]{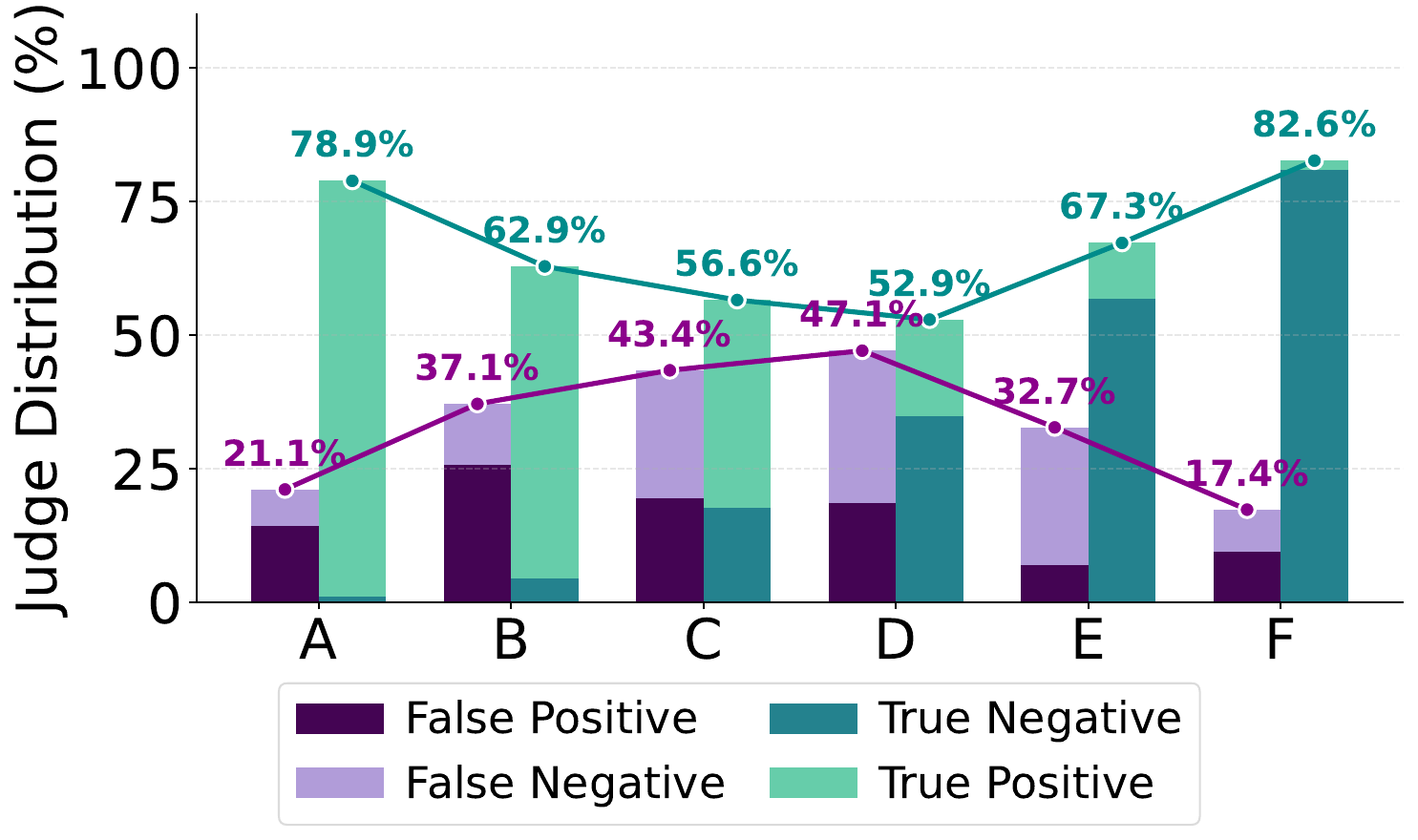}
		\caption{Coder-32B}
		\label{fig:ds_v3_etc_cross}
	\end{subfigure}
    \hfill
    \begin{subfigure}{0.49\linewidth}
        \includegraphics[width=\linewidth]{pdf/ds_v3_cte.pdf}
        \caption{DeepSeek-V3}
        \label{fig:qwq_32b_etc_cross}
    \end{subfigure}
    \hfill
    \begin{subfigure}{0.49\linewidth}
        \includegraphics[width=\linewidth]{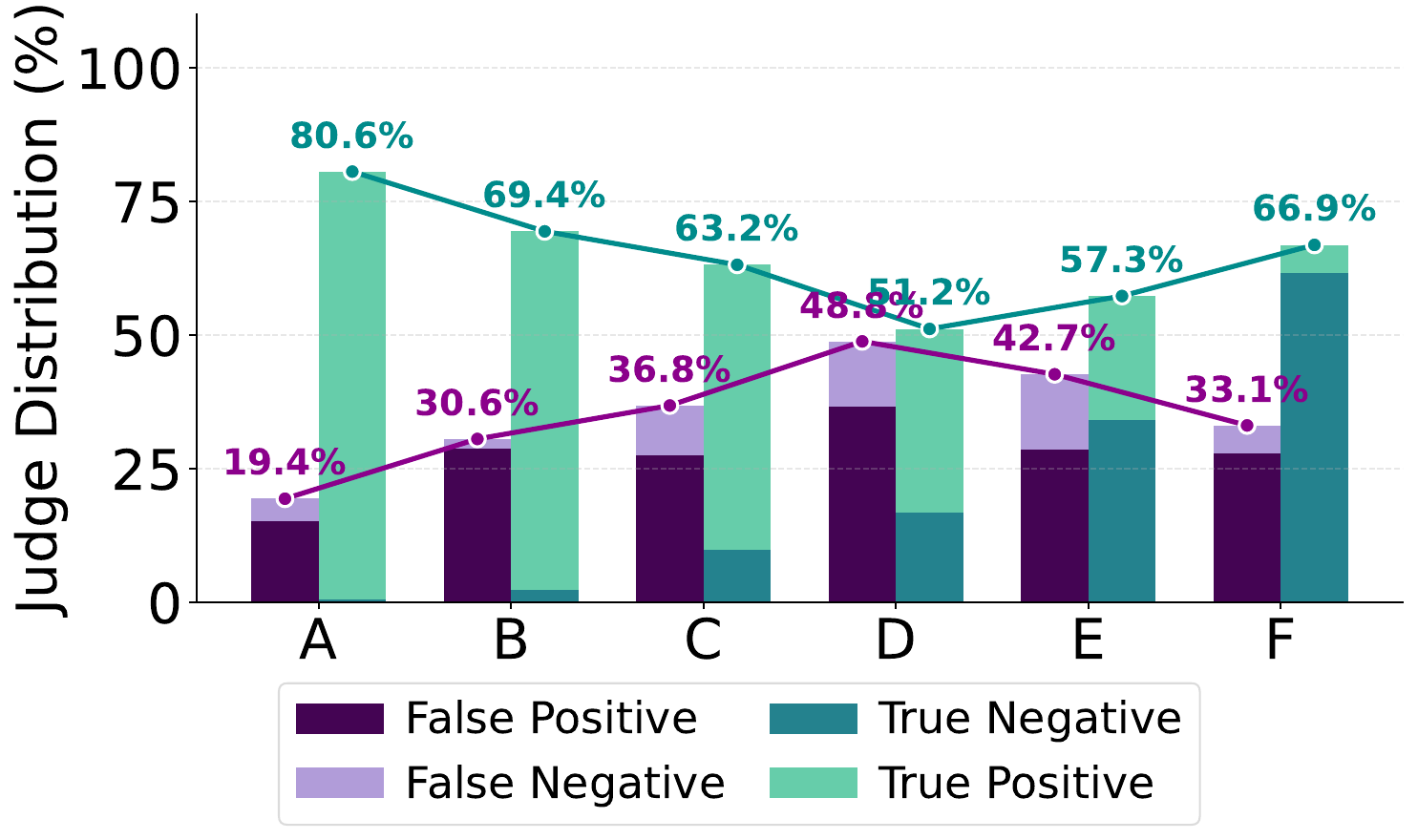}
        \caption{QwQ-32B}
        \label{fig:qwq_32b_etc_cross}
    \end{subfigure}
	\caption{Judge distribution of DeepSeek-V3 over solutions from different models.}
	\label{fig:cte_cross}
\end{figure*}

We further conduct experiments to investigate whether LLMs can identify mistakes made by other models, aiming to explore the judging capability of LLMs. Specifically, we utilize DeepSeek-V3 to determine whether the solutions generated by other models contain errors. As shown in Figure \ref{fig:cte_cross}, the model is also capable of identifying mistakes in the solution, and the accuracy all exhibits a decreasing-then-increasing trend.

For relatively weaker models (such as coder-32B and Qwen-72B), which lack the ability to solve difficult problems, DeepSeek-V3 can easily identify the errors in their solutions to such challenging problems, which is reflected by True Negatives constituting the majority of accurate judgments.
Moreover, in the cases where DeepSeek-V3 judges its own solutions and those of QWQ-32B, we observe that DeepSeek-V3 shows a stronger tendency to judge its own solutions as correct, revealing a form of self-consistency in the model's self-perception.

\subsection{How does case generation benefit from model mixture?}
\begin{figure*}[!h]
	\centering
	\begin{subfigure}{0.3\linewidth}
		\includegraphics[width=\linewidth]{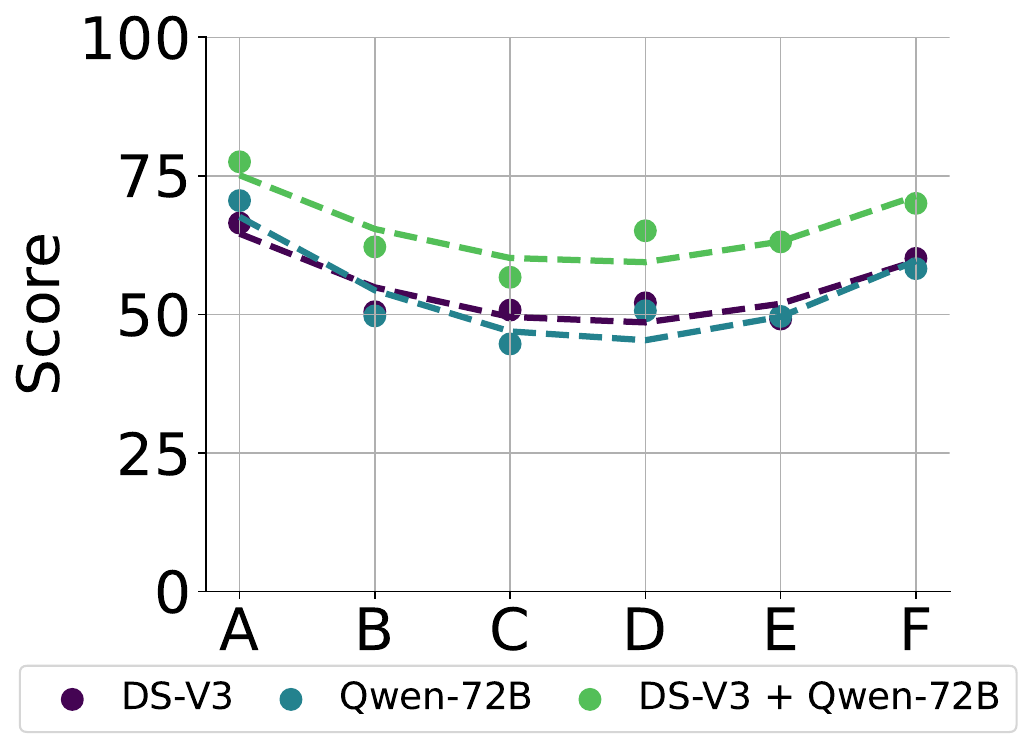}
		\caption{Qwen2.5-72B-Instruct}
		\label{fig:qwen_72b_etc_cross}
	\end{subfigure}
    \hfill
	\begin{subfigure}{0.3\linewidth}
		\includegraphics[width=\linewidth]{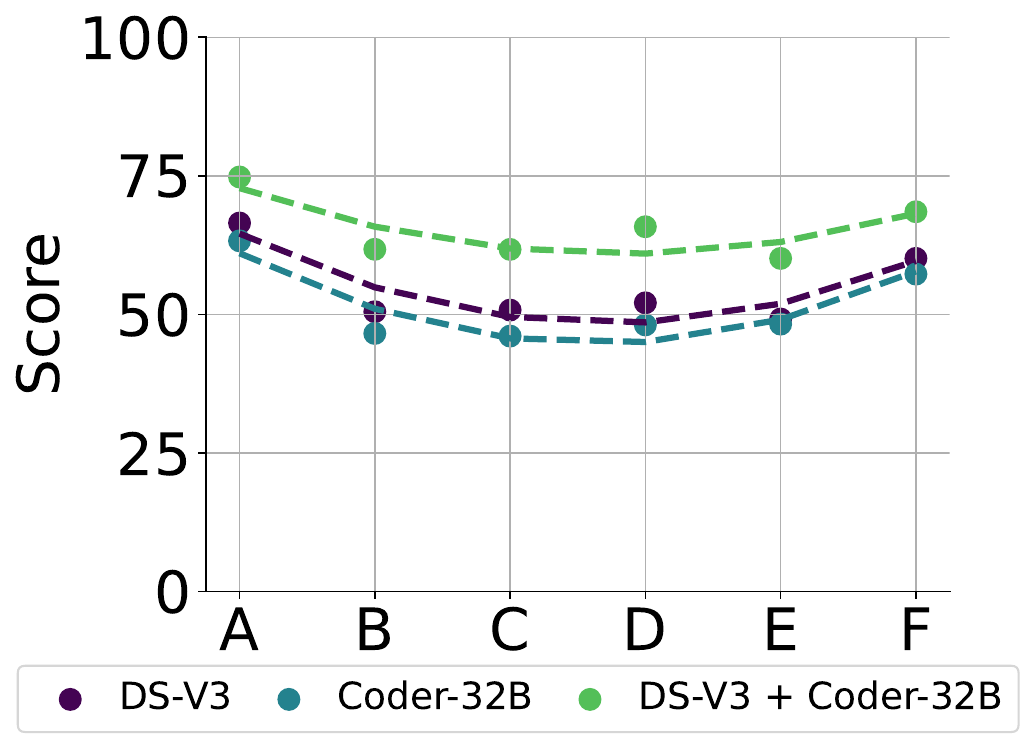}
		\caption{Coder-32B-Instruct}
		\label{fig:ds_v3_etc_cross}
	\end{subfigure}
    \hfill
    \begin{subfigure}{0.3\linewidth}
        \includegraphics[width=\linewidth]{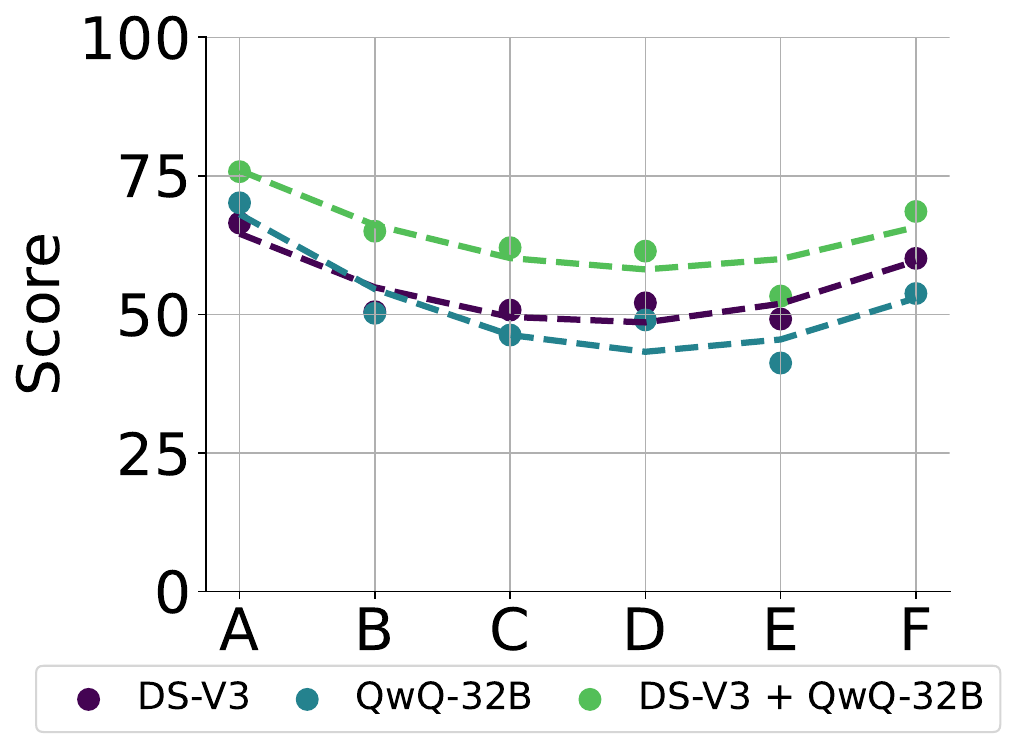}
        \caption{QwQ-32B}
        \label{fig:qwq_32b_etc_cross}
    \end{subfigure}
	\caption{Case score of model mixture between DeepSeek-V3 and other models.}
\label{fig:mix_appendix}
\end{figure*}

To further validate the effectiveness of model mixture, we merge the cases generated by different models and calculate the accuracy of these case sets. The results are presented in Figure \ref{fig:mix_appendix}. We observe that the model mixture brings significant improvements in case score performance, as different models exhibit different biases and the diversity introduced by these models in the case generation task directly contributes to the improvement in scores.

\begin{figure*}
\begin{tcolorbox}[
colback=white!10!white,
colframe=black!50!white,
title=Editorial Generation Prompt,
]
\small
[[ Instruction ]]  

You are a code expert, and your goal is to develop a editorial analysis for a given coding problem. You will analyze the problem, explain the approach, including any necessary constraints and mathematical formulas, to help the reader understand how to solve it. Your explanation should not include the final code but should guide the reader to implement it themselves. Please indicate the time complexity at the end of the analysis.

[[ Constraints ]]  

Your editorial analysis must satisfy the following:  

- Clearly explain the approach to solve the problem.  

- Include any necessary constraints from the problem statement.  

- Use mathematical formulas where applicable to clarify the solution.  

- Do not provide the final code, only the editorial.  

- Indicate the time complexity of the solution at the end.  

- Ensure that the editorial addresses all aspects of the problem as stated.

[[ Example ]]  

[Editorial]

The leaves of $T$ are leaves of initial stars, and those vertices distant by at most two from it belong to the same star. Thus, the following algorithm is possible.

- Choose a leaf of $T$.  

- Count the number of vertices distant by at most two from it (including itself). If there are $x$ of them, they form a level-$(x-1)$ star.  

- Remove the counted vertices and adjacent edges from $T$.

After you repeat this until no vertices remain in $T$, you obtain the answer. With an appropriate implementation, it works in a total of $\mathrm{O}(N)$ time, but the implementation is a bit complicated. We describe a simpler implementation with some observations.

The following lemma holds:  

- In an original star, let us call the non-leaf vertex the center. In $T$, the distance between centers is always a multiple of $3$, and that between a center and a leaf is always a non-multiple of $3$.

This can be shown by induction. With this lemma, we can come up with the following algorithm:  

- Choose a vertex from $T$ that was the center of an original star. You can do so by choosing the vertex adjacent to a leaf.  

- Find the shortest distance of each vertex from the chosen vertex.  

- For each vertex whose shortest distance is a multiple of $3$, add the degree of that vertex to $L$.

[Time Complexity] $\mathrm{O}(N)$

[[ Output Format ]]

Your response should be formatted as follows and should not include any additional information:

[Think] Your thinking about the reasoning process in the mind.

[Editorial] Your final editoral of the problem.

[Time Complexity] The time complexity of your solution.

[[ Problem Begin ]]

\red{\{problem\}}

[[ Problem End ]]

\label{fig:critique-prompt}
\end{tcolorbox}
\end{figure*}

\begin{figure*}
\begin{tcolorbox}[
colback=white!10!white,
colframe=black!50!white,
title=Solution Generation Prompt,
]
\small
[[ Instruction ]]  

You are an expert C++ programmer. Your goal is to generate a complete, correct C++ program for a given coding problem. The program should handle all edge cases, follow best practices, and be efficient where necessary. Enclose your program within C++ code delimiters as shown below.

[[ Constraints ]]  

- Generate a fully functional C++ solution that compiles and passes all tests.  

- The code should be standalone and not rely on external libraries beyond what's standard in C++, unless specified in the problem.  

- Adhere strictly to the problem's input and output formats.  

- Ensure the code is clean, well-indented, and includes comments to explain complex logic.  

- Include all necessary headers and use the standard namespace.  

- Wrap your code in triple backticks with C++ annotation.

[[ Example ]]

\#include <bits/stdc++.h>

using namespace std;

int main() \{

    // sample solution

    return 0;

\}

[[ Output Format ]]

Your response should be formatted as follows and should not include any additional information:

[Analysis] Your analysis of the problem.

[Code] Your C++ code in a code block.

[[Problem begin]]

\red{\{problem\}}

[[Problem end]]

\label{fig:critique-prompt}
\end{tcolorbox}
\end{figure*}

\begin{figure*}
\begin{tcolorbox}[
colback=white!10!white,
colframe=black!50!white,
title=Case Generation Prompt,
]
\small

[[ Instruction ]]

You are an expert Python competitive programmer and your goal is to construct input-generators for testing programming contest problems. You will write relevant generators and finally implement a `construct\_inputs` function that returns a list of 50 diverse inputs sampled from those generators. Remember to strictly follow the instructions and constraints present in the problem statement.

[[ Constraints ]]

Your input-generators and `construct\_inputs` must satisfy all of the following:

  - **Deterministic framework**: the code may call randomness internally, but the overall scheme and parameter ranges must be hard-coded (no external configuration or user prompts).

  - **Coverage**: include edge-case ranges (smallest/largest sizes, boundary weight values), typical scenarios, and stress scenarios near the problem’s limits.

  - **Validity**: generated inputs must always respect the problem’s stated input format and numeric bounds (e.g. $1 \le N \le N_{max}, weight_{min} \le weight_i \le weight_{max}$, etc.).

  - **Reproducibility**: allow for seeding if needed (e.g. accept a seed parameter), but default behavior needs no external input.

  - **Diversity**: return a list containing at least three tiers of size/scale (e.g. small, medium, large) and within each tier cover multiple parameter combinations.

  - **Clarity**: each testcase’s `input` string must be parseable by the contestant’s code.

[[ Example ]]
\begin{lstlisting}[language=Python]
import numpy as np
def random_input_generator(weight_min, weight_max, size_min, size_max, seed=None):
    if seed is not None:
        np.random.seed(seed)
    n = np.random.randint(size_min, size_max+1)
    weights = np.random.randint(weight_min, weight_max+1, size=n).tolist()
    k = np.random.randint(1, n+1)
    return { 'input': ' '.join(map(str, weights)) + ' ' + str(k) + '\\n' }

def edge_case_generator(case_id):
    cases = [
        # 0: smallest size, smallest weight
        { 'input': '1 1\\n' },
        # 1: smallest size, largest weight
        { 'input': '1 1000000000\\n' },
        ...
        # 9: mixed boundary in medium
        { 'input': '1000 ' + ' '.join(['1']*499 + ['1000000']*500) + ' 250000\\n' },
    ]
    return cases[case_id]

def construct_inputs():
    inputs_list = []
    # 10 edge cases
    for i in range(10):
        inputs_list.append(edge_case_generator(i))
    # small tier
    for i in range(10, 20):
        inputs_list.append(random_input_generator(1, 10**3, 1, 10, seed=i))
    # medium tier
    for i in range(20, 35):
        inputs_list.append(random_input_generator(1, 10**6, 1, 10**3, seed=i))
    # large tier
    for i in range(35, 50):
        inputs_list.append(random_input_generator(1, 10**9, 1, 10**5, seed=i+100))

    return inputs_list
\end{lstlisting}

[[ Output Format ]]

Your response should be formatted as follows and should not include any additional information:

[Analysis] Your analysis of the problem.

[Code] Your Python scripts in a code block.

[[ Problem Begin ]]

\red{\{problem\}}

[[ Problem End ]]

\label{fig:critique-prompt}
\end{tcolorbox}
\end{figure*}

\begin{figure*}
\begin{tcolorbox}[
colback=white!10!white,
colframe=black!50!white,
title=Editorial Judge Prompt,
]
\small
[[ Instruction ]]  

You are a code expert and judge. Your goal is to evaluate a candidate’s editorial for a given coding problem, using the official editorial as a reference. First, extract the time complexities from both the official editorial and the candidate’s editorial. If the candidate’s time complexity is asymptotically worse than the official one, assign a score of 0. Otherwise, analyze whether the candidate’s solution is logically correct and solves the problem as required. Note that the candidate’s approach may differ from the official one, but it should still be a valid solution to the problem. Finally, assign a binary score: 1 if the solution is correct and has an acceptable time complexity, otherwise 0.

[[ Constraints ]]  

Your judgment must satisfy the following:  

- Read and understand the problem statement in full.  

- Use the official editorial as a reference to verify the correctness of the candidate’s approach, but allow for different valid solutions.  

- Assign a score of 1 if the candidate’s solution is logically correct and has an acceptable time complexity, otherwise assign 0.

[[ Output Format ]]  

Your response must follow exactly this format without any extra information:  

[Analysis] Your analysis of the problem, the two editorials, and the correctness of the candidate’s solution.  

[Score] Your final judge score.

[[ Problem Begin ]]

\red{\{problem\}}

[[ Problem End ]]

[[ Official Editorial Begin ]]

\red{\{gt editorial\}}

[[ Official Editorial End ]]

[[ Candidate Editorial Begin ]]

\red{\{editorial\}}

[[ Candidate Editorial End ]]

\end{tcolorbox}
\end{figure*}

\begin{figure*}
\begin{tcolorbox}[
colback=white!10!white,
colframe=black!50!white,
title=Solution Judge Prompt,
]
\small
[[ Instruction ]]  

You are a programming competition judge. Your task is to analyze a submitted solution for a specified problem and determine its correctness. You should focus on logical correctness, coverage of all edge cases, and any implementation flaws that would cause test failures.

[[ Constraints ]]  

- Provide a detailed analysis of potential logical errors or omissions.  

- Indicate whether the solution passes all test cases or fails some.  

- Do not execute code; base your judgment on static reasoning.   

- Assign a score of 1 if the candidate’s solution is logically correct and has an acceptable time complexity, otherwise assign 0.

[[ Example ]]  

[Analysis] The solution attempts binary search but has an off-by-one error in the termination condition (line 8). This causes incorrect results when the target is at array boundaries.

[Score] 0

[[ Output Format ]]

Your response should be formatted as follows and should not include any additional information:

[Analysis] Your analysis of the problem and the solution.

[Score] Your 0/1 score of the solution.

[[ Problem begin ]]

\red{\{problem\}}

[[ Problem end ]]

[[ Solution begin ]]

\red{\{solution\}}

[[ Solution end ]]

\end{tcolorbox}
\end{figure*}

\begin{figure*}
\begin{tcolorbox}[
colback=white!10!white,
colframe=black!50!white,
title=Case Judge Prompt,
]
\small
[[ Instruction ]]

You are a programming competition judge. Your task is to determine whether a given test case’s input and output match according to the problem’s specification. Focus solely on whether the provided output is the correct result for the provided input under the problem logic.

[[ Constraints ]]

* Check that the “Case Output” is exactly what the problem would produce for the given “Case Input.”

* Identify any discrepancies, incorrect results, or mismatches.

* Do not assess solution code—only compare input versus output.

* Do not execute code; base your judgment on logical reasoning and the problem statement.

* Assign a score of 1 if the candidate’s case is logically correct, otherwise assign 0.

[[ Example ]]

[Analysis] For input `3 5 2`, the problem asks for the sum of the first two numbers. The expected result is `8`, but the provided output is `15`, so they do not match.

[Score] 0

[[ Output Format ]]

Your response must follow this exact format, with no additional text:

[Analysis] <your detailed analysis of input-output matching>

[Score] <0 or 1>

[[ Problem begin ]]

\red{\{problem\}}

[[ Problem end ]]

[[ Case Input Begin]]

\red{\{case input\}}

[[ Case Input end ]]

[[ Case Output Begin ]]

\red{\{case output\}}

[[ Case Output end ]]

\label{fig:critique-prompt}
\end{tcolorbox}
\end{figure*}

\end{document}